\documentclass[10pt,journal,compsoc]{IEEEtran}

\ifCLASSOPTIONcompsoc
  \usepackage[nocompress]{cite}
\else
  \usepackage{cite}
\fi

\usepackage{amsmath,amsfonts,amsthm,amssymb}
\usepackage{algorithmic}
\usepackage{algorithm}
\usepackage{array}
\usepackage[caption=false,font=scriptsize,labelfont=rm,textfont=rm]{subfig}
\usepackage{textcomp}
\usepackage{stfloats}
\usepackage{url}
\usepackage{verbatim}
\usepackage{graphicx}
\usepackage{mdwmath}
\usepackage{mdwtab}
\usepackage{eqparbox}
\usepackage{multirow}
\usepackage{enumitem}
\usepackage[colorlinks=true,linkcolor=black,citecolor=blue,urlcolor=blue,]{hyperref}

\renewcommand{\algorithmicrequire}{\textbf{Input:}}  
\renewcommand{\algorithmicensure}{\textbf{Output:}}  
\renewcommand{\algorithmicprint}{\textbf{break}}     
\newcommand{\eg}{e.g.}
\newcommand{\ie}{{i.e.}}
\newcommand{\etal}{\textit{et~al.\ }}

\graphicspath{{Figure/}}

\begin{document}

\title{CylinderTag: An Accurate and Flexible Marker \\ for Cylinder-Shape Objects Pose Estimation \\ Based on Projective Invariants}

\author{Shaoan~Wang,
        Mingzhu~Zhu,
        Yaoqing~Hu,
        Dongyue~Li,
        Fusong~Yuan,
       and Junzhi~Yu,~\IEEEmembership{Fellow,~IEEE}
        \thanks{S.~Wang, Y.~Hu, D.~Li, and J.~Yu are with the State Key Laboratory for Turbulence and Complex Systems, Department of Advanced Manufacturing and Robotics, College of Engineering, Peking University, Beijing 100871, China (e-mail: wangshaoan@stu.pku.edu.cn; 2101111894@stu.pku.edu.cn; 2001111648@stu.pku.edu.cn; junzhi.yu@ia.ac.cn).}
        \thanks{M.~Zhu is with the Department of Mechanical Engineering, Fuzhou University, Fuzhou 350000, China (e-mail: mzz@fzu.edu.cn).}
        \thanks{F.~Yuan is with the National Engineering Laboratory for Digital and Material Technology of Stomatology, Center of Digital Dentistry, Peking University School and Hospital of Stomatology, Beijing 100190, China (e-mail: yuanfusong@bjmu.edu.cn).}
        \thanks{This work was supported in part by the National Key Research and Development Program of China under Grant 2020YFB1312800, in part by the National Natural Science Foundation of China under Grant 62003007, and in part by the Peking University Clinical Scientist Training Program under Grant BMU2023PYJH019. (Corresponding author: Junzhi Yu.)}
}

\markboth{IEEE Transactions on Visualization and Computer Graphics}
{Wang \MakeLowercase{\textit{et al.}}: CylinderTag: An Accurate and Flexible Marker for Cylinder-Shape Objects Pose Estimation Based on Projective Invariants}

\IEEEtitleabstractindextext{
\begin{abstract}
High-precision pose estimation based on visual markers has been a thriving research topic in the field of computer vision. However, the suitability of traditional flat markers on curved objects is limited due to the diverse shapes of curved surfaces, which hinders the development of high-precision pose estimation for curved objects. Therefore, this paper proposes a novel visual marker called CylinderTag, which is designed for developable curved surfaces such as cylindrical surfaces. CylinderTag is a cyclic marker that can be firmly attached to objects with a cylindrical shape. Leveraging the manifold assumption, the cross-ratio in projective invariance is utilized for encoding in the direction of zero curvature on the surface. Additionally, to facilitate the usage of CylinderTag, we propose a heuristic search-based marker generator and a high-performance recognizer as well. Moreover, an all-encompassing evaluation of CylinderTag properties is conducted by means of extensive experimentation, covering detection rate, detection speed, dictionary size, localization jitter, and pose estimation accuracy. CylinderTag showcases superior detection performance from varying view angles in comparison to traditional visual markers, accompanied by higher localization accuracy. Furthermore, CylinderTag boasts real-time detection capability and an extensive marker dictionary, offering enhanced versatility and practicality in a wide range of applications. Experimental results demonstrate that the CylinderTag is a highly promising visual marker for use on cylindrical-like surfaces, thus offering important guidance for future research on high-precision visual localization of cylinder-shaped objects. The code is available at: \url{https://github.com/wsakobe/CylinderTag}.
\end{abstract}

}

\maketitle
\IEEEpeerreviewmaketitle

\ifCLASSOPTIONcompsoc
\IEEEraisesectionheading{\section{Introduction}}
\else
\section{Introduction}
\fi
\IEEEPARstart{P}{ose} estimation plays a crucial role in the vision-driven interacting tasks ranging from VR/AR \cite{Bib:Chen2022TVCG}, human-robot interaction (HRI) \cite{Bib:Du2021AIR}, to motion capture (MoCap) \cite{Bib:Lv2018TVCG}. As more advanced applications emerge, there is a growing demand for more accurate, convenient, and environment-adaptive 6-DoF object pose estimators. Traditional methods focus on designing fiducial markers with encoded code within the pattern on the marker, enabling the recognition algorithm to detect the pattern and decode the unique information stored in different markers. These markers are rigidly mounted to objects whose poses are to be estimated and are used to provide high-precision features (\eg, corners, lines, and circles) to estimate the relative 6-DoF poses between the objects frame and the camera frame.

With the continuous development of advanced technology, the demand for high-precision pose estimation is increasing in complex scenarios, particularly those involving curved objects. Square markers are the most widely used type of visual marker, but their recognition algorithm depends on the structural properties of the square, which can result in recognition failure due to surface deformation. As a result, these markers can only be mounted with low adhesion on curved objects, making them difficult to decode from most viewing angles. Circular markers can usually withstand a certain degree of deformation, which means they can be used on surfaces with some curvature, however, their localization relies on the fitting of circular features, and surface deformation can lead to the deformation of circular features, which can make localization inaccurate. Topological markers, which do not require canonical modeling, can mitigate the effect of surface deformation. However, they are not cyclic and require a limited viewpoint for full recognition.

Several recently developed marker systems have recognized the effects of deformation and are utilizing emerging technologies, such as deep learning, to address this issue. Through end-to-end training, a large number of images simulating marker deformation can be generated, which enhances the detection performance in the presence of deformation. Despite the significant improvement in recognition performance for deformation, existing deep learning-based marker systems are still limited to square shapes and do not account for the nature of the deformed surface, resulting in low localization accuracy.

In this work, a fiducial marker system towards the cylindrical-like surfaces called CylinderTag is proposed. CylinderTag incorporates a long strip feature, which is based on the manifold assumption of cylindrical-like surfaces, and employs the cross-ratio as the encoding method. We propose both the marker generator and recognizer, and conduct experiments to verify the advantages of this marker in recognition and localization compared to traditional markers. The main contributions of this work are elaborated as follows:
\begin{enumerate}[leftmargin=15pt]
    \item A novel visual marker system called CylinderTag is proposed for cylindrical-like surface design, with an encoding principle based on projective invariants.
    \item A CylinderTag marker generator based on heuristic search is proposed to efficiently generate a sufficient number of markers.
    \item A real-time CylinderTag marker recognizer is proposed. Extensive experiments are conducted to verify its advantages in detection and localization accuracy.
\end{enumerate}

\begin{figure*}[!t]
\centering
\includegraphics{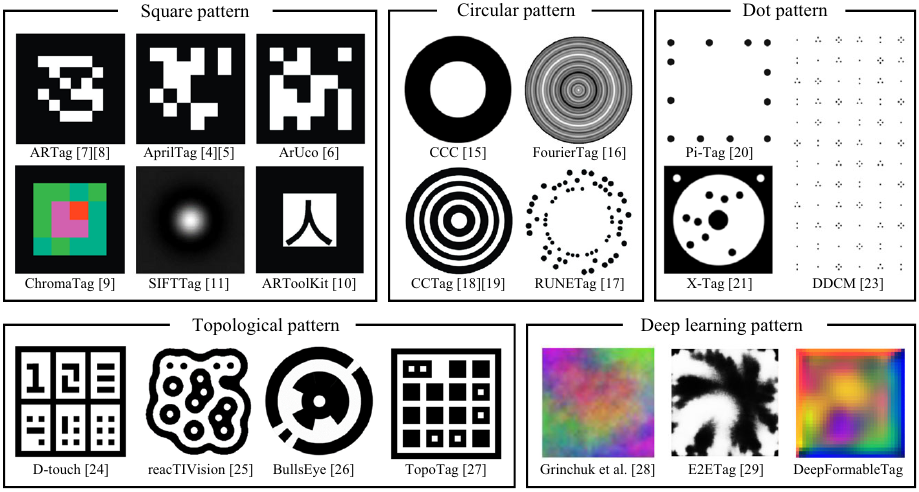}
\caption{Various types of visual markers, classified into square, circular, dot, topological, and deep learning patterns.}
\label{Fig:existmarker}
\end{figure*}

The remainder of this paper is organized as follows. Section~2 describes the literature survey on existing marker systems. Section~3 provides a detailed explanation of the CylinderTag system, which comprises the marker generator and recognizer. Several experiments were conducted in Section~4 to evaluate the design, detection, localization jitter, and pose estimation performance of CylinderTag compared to traditional markers, as well as to showcase its versatility in various application scenarios. Section~5 provides the conclusions and future work of this paper.

\section{Related Works}
Since the invention and widespread use of 2D barcodes such as Data Matrix, Maxi code, and QR code in the late $1980$s, more research has focused on developing more generalized, robust, and stable fiducial markers. Nowadays, various fiducial markers with different applications or advantages have been developed, as shown in Fig.~\ref{Fig:existmarker}. These markers can be categorized as follows:

\textit{Square pattern} The square pattern usually comprises a high-contrast black square border and an internal coded pattern. AprilTag \cite{Bib:Wang2016IROS,Bib:Maximilian2019IROS}, ArUco \cite{Bib:Garrido2014PR}, and ARTag\cite{Bib:Fiala2010TPAMI,Bib:Fiala2005CVPR} are the most commonly used marker systems in robot navigation and AR/VR today. Their internal patterns are similar by creating tight black/white regions to encode binary codes. In addition, for system robustness, the Cyclic Redundancy Check (CRC) or Reed-Solomon codes which are commonly used in the communication field are introduced to improve the capability of the error correlation mechanism. ChromaTag \cite{Bib:DeGol2017ICCV} introduces LAB colorspace that replaces black/white region into different colors over AprilTag to improve marker detection speed. ARToolKit \cite{Bib:Wagner2007CVWW} uses image cross-correlation to obtain marker information, which means the internal pattern can be arbitrary except for symmetric ones. Tanaka \etal designed LentiMark \cite{Bib:Tanaka2012ICRA} and its improved version \cite{Bib:Tanaka2017IROS} based on VMP and FDP in order to solve the problem of pose ambiguity of planar markers. Tateno \etal designed Nested Marker \cite{Bib:Tateno2007VRC} by introducing a recursive layered structure and thus realizing robust marker recognition at different scales. Schweiger \etal \cite{Bib:Schweiger2009VMVW} proposed markers that are specifically designed for scale-invariant feature detectors SIFT and SURF. These two detectors can obtain the highest response scores, resulting in very high detection rates at different scales. Herout \etal \cite{Bib:Herout2013CVPR} used De Bruijn tori for information encoding, which means that only a subregion on the marker needs to be observed to decode the corresponding information. Garrido-Jurado \etal \cite{Bib:Garrido2016PR} also focused on marker generation tasks, using mixed integer programming to accelerate the generation of ArUco. Zhu \etal \cite{Bib:Zhu2022PAMI,Bib:Wang2022TIM} proposed the binary wave function collapse (bWFC) algorithm that can arbitrarily set the shape of the pattern and the subregion with a fast-generating speed.

\textit{Circular pattern} The circular pattern has also received significant attention due to the property that the quadratic form remains invariant under projection transformation, which means that the circle can remain quadratic in most views. This property helps the detector to easily segment out the marked area. In the earliest work, Concentric Contrast Circles (CCC) \cite{Bib:Gatrell1992CIRS} was first introduced in target design with a circular pattern. Subsequent work was based on this by using multiple rings of different colors. In addition, FourierTag \cite{Bib:Xu2011CRV} utilizes frequency images as features for adequate identification. RUNETag \cite{Bib:Bergamasco2016TPAMI} arranges the points uniformly into concentric circles and introduces a generalized BCH code for the strongest error correction and anti-obscuration capability. CCTag \cite{Bib:Calvet2012ICIP,Bib:Calvet2016CVPR} is an improvement of CCC. Its main contribution is the robustness of motion blur.

\textit{Dot pattern} Because of the excellent noise resistance of the dot feature, patterns with dots also raise some attention. Pi-Tag \cite{Bib:Bergamasco2013MVA} is composed of equal-sized dots distributed around the edges of the square. It adopted cross-ratio, which is one of the projective invariants, as the encoder. Consisting of multiple dots randomly distributed in the middle of two concentric circles, X-Tag \cite{Bib:Birdal20163DV} similarly utilized cross-ratios and intersection preservation constraints as a reference for decoding. In addition, Watanabe \etal \cite{Bib:Watanabe2017ISMAR} proposed Extended Dot Cluster Marker (EDCM), which is an improved version of Deformable Dot Cluster Marker (DDCM) \cite{Bib:Narita2016TVCG}. These two markers are encoded by using the number of dots in small areas as features to obtain a strong resistance to occlusion.

\textit{Topological pattern} Another widely developed category of the marker is encoded based on the topological information, which demonstrates the ability to improve robustness against noise and brightness. ReacTIVision \cite{Bib:Kaltenbrunner2007TEI} provided unique identities purely with the topological structure by building a left heavy depth sequence of the region adjacency graph. BullsEye \cite{Bib:Klokmose2014ICITS} consists of a central white dot surrounded by a solid black ring and one or more data rings again surrounded by a solid white ring inside a black ring with three white studs. Although this type of marker can be decoded without the limitation of shape, its localization performance is also affected by the lack of sufficient features. TopoTag \cite{Bib:Yu2020TVCG} utilized the structure of the square pattern and only used the topological pattern as the encoder, which offered enough corner features to locate.

\textit{Deep learning pattern} With the rapid development of deep learning, many scholars have pondered whether deep neural networks can be used instead of humans to design more robust and stable markers. A feasible solution was first proposed by Grinchuk \etal \cite{Bib:Grinchuk2016NIPS}, which used a synthesizer network to generate patterns that are difficult for humans to understand. E2ETag \cite{Bib:Peace2020BMVC} is an attempt to build the first end-to-end trainable fiducial marker generator, while the very small size of the code dictionary and the slow detecting speed both limit its wide application. Recently, DeepFormableTag \cite{Bib:Yaldiz2021TOG} brought sufficient attention to this field. By proposing a differentiable image simulator that can render images including close-to-realistic reflection, color alteration, imaging noise, and shape deformation of markers, a large image dataset containing the majority of real-world situations is generated, resulting in the most robust detector to date. Currently, using deep learning to generate new types of markers is still in the preliminary stage, and although it has advantages that are difficult to compare with artificial markers, it is still limited to the traditional square marker shape.

\section{CylinderTag System}
CylinderTag is dedicated to designing a marker system with no additional accessories (like the infrared sphere-based optical MoCap system \cite{Bib:Pan2022TIE}) and uniformity in any viewing angle for high-precision pose estimation applications of cylindrical objects. For markers that are tightly attached to cylindrical objects, a problem that needs to be overcome is the deformation of the surface perpendicular to the cylindrical axis. However, the planes passing through the cylindrical axis intersect the cylindrical surface as lines, revealing that the cross-ratio in these lines satisfies the perspective invariants. Furthermore, under the manifold assumption, any narrow enough strip of surface in a cylinder parallel to the direction of the cylindrical axis can be regarded as a plane, providing ideas for the design of the features in CylinderTag.

\subsection{Marker Design \& Generator}
\begin{figure}[!t]
\centering
\includegraphics[width=.35\textwidth]{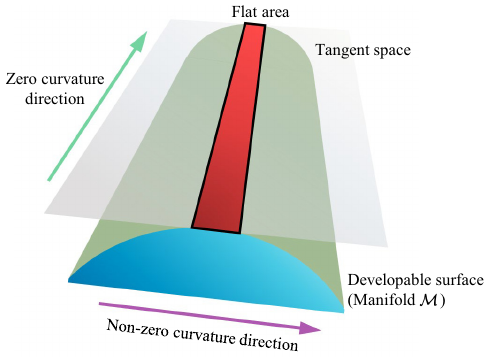}
\caption{\textbf{Schematic diagram of the manifold assumption.} The neighborhood of a point on the manifold can be approximated by its tangent space.}
\label{Fig:manifold}
\end{figure}

In this subsection, the design principle and basic structure of CylinderTag are introduced in detail. Based on simulation experiments, the nature of the cross-ratio as a projection invariant is analyzed, and the design of CylinderTag is guided by the analysis results. In addition, a CylinderTag generator based on heuristic search is proposed, which can heuristically select the optimal solution according to the potential probabilities of different solutions in the current solution space, thus significantly improving the speed and efficiency of marker generation.

\subsubsection{Manifold Assumption}
For developable surfaces, such as cylindrical and conical surfaces, the property of zero curvature in a given direction is satisfied. That is, a straight line in the direction of zero curvature can be kept straight. Therefore, some projective invariants associated with lines can be considered to participate in the design of the marker. Cross-ratio (cr) is one of the most common projective invariants in projective geometry and is also widely used for feature encoding of visual markers \cite{Bib:Tsonisp1998IROS,Bib:Liere2003VR,Bib:Coutinho2012IJCV}. It describes the proportional invariance of four points that are invariant in a line under a projective transformation. This property can be well applied to the zero curvature direction of the developable surface described above. For the other non-zero curvature direction, it can be conceptualized as a manifold $\mathcal{M}$ in $\mathbb{R}^3$ where the local neighborhood of $\mathcal{M}$ can be treated as a tangent space of $\mathcal{M}$ and satisfying the planar property, as shown in Fig.~\ref{Fig:manifold}.

\subsubsection{The Nature of Cross-ratio}
\begin{figure}[!t]
\centering
\includegraphics[width=.35\textwidth]{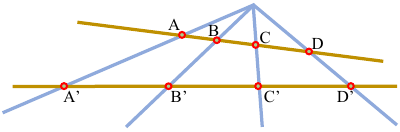}
\caption{\textbf{Schematic diagram of the cross-ratio.} Under the projective transformation, the cross-ratios corresponding to the four points on the same line remain unchanged.}
\label{Fig:crossratio}
\end{figure}

For four fixed points $A, B, C, D$ located on a straight line and four points $A', B', C', D'$ under projective transformation, as shown in Fig.~\ref{Fig:crossratio}, the cross-ratio is formulated as:
\begin{equation}
(A, B ; C, D)=(A', B' ; C', D')=\frac{A C}{C B} / \frac{A D}{D B}
\end{equation}

\begin{figure}[!t]
\centering
\includegraphics[width=.4\textwidth]{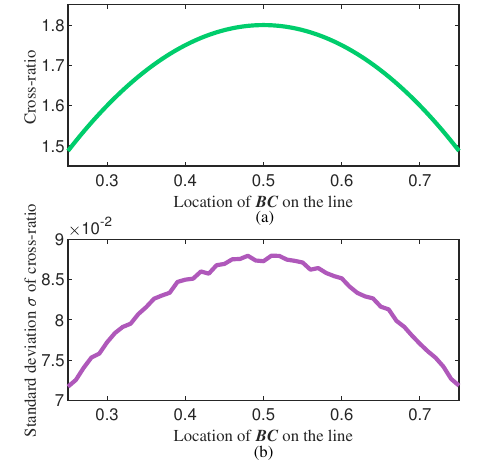}
\caption{\textbf{The nature of the cross-ratio.} (a) The value of the cross-ratio with respect to the location of $BC$. (b) The standard deviation of the cross-ratio with respect to the location of $BC$.}
\label{Fig:crnature}
\end{figure}

In order to make better use of the cross-ratio, the properties it holds deserve to be studied and analyzed. The nonlinear nature of the cross-ratios is often discussed \cite{Bib:Huynh2000BMVC}, and as Fig.~\ref{Fig:crnature}a shows, the cross-ratios change more slowly near the midpoint. Next, if the cross-ratio is used as the encoding principle for CylinderTag, it is also necessary to analyze the error of the cross-ratio in different cases. Here, we perform simulation experiments for the general case. For four points $A, B, C, D$ on a two-dimensional straight line of fixed length, we fix the positions of points $A$ and $D$ and fix the distance $BC$. An i.i.d. 2D Gaussian noise is applied to the positions of these points to simulate the corner localization errors. Calculate the standard deviation $\sigma$ of the cross-ratio calculation when the midpoint of $BC$ is located at different positions on the line. As illustrated in Fig.~\ref{Fig:crnature}b, the standard deviation of the cross-ratio under Gaussian noise exhibits a trend that bears resemblance to that of the cross-ratio itself. Notably, the standard deviation is significantly higher near the midpoint compared to the ends. This indicates that under the same noise level, the standard deviation of the cross-ratio near the midpoint is larger, while the derivative of the cross-ratio near the midpoint is smaller, which together lead to a lower decoding success rate in this region. Hence, the regions closer to the ends are chosen as the encoding regions, and the regions near the midpoints are ignored, thus guaranteeing the decoding performance.

\begin{figure}[!t]
	\centering
	\subfloat[]{\includegraphics[width=.35\textwidth]{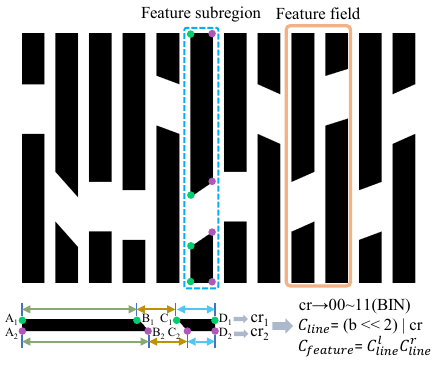}\label{Fig:CylinderTag}}
		\vspace{-8pt}
	\subfloat[]{\includegraphics[width=.35\textwidth]{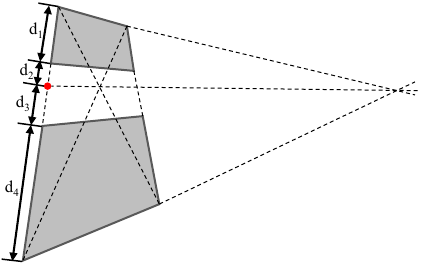}\label{Fig:position}}
	\caption{\textbf{The structure of CylinderTag.} (a) Schematic diagram of the pattern and encoding principle of CylinderTag (12c2f). (b) Schematic diagram of the method to distinguish the symmetry of the cross-ratio.}
    \label{Fig:structure}
\end{figure}

\subsubsection{The Structure of CylinderTag}
CylinderTag is composed of several ``feature subregions''. Each ``feature subregion'' consists of two co-elongated quads with equal lengths of short edges. Fig.~\ref{Fig:CylinderTag} illustrates the structure of CylinderTag (12c2f), where ``12c'' means the number of the columns (feature subregions) and ``2f'' means the length of the feature field. For each ``feature subregion'', we assume that the direction of the long edges aligns with the direction of zero curvature on the surface. Intuitively, in this case, the four corners corresponding to the long edges remain co-linear in different poses, and their corresponding cross-ratio should also remain constant. We categorize the cross-ratio as the ID of each ``feature subregion'', which can be known by detecting the corner positions on each long side and computing their corresponding cross-ratio. The cross-ratio of a ``feature subregion'' is thus determined by the lengths of the long sides of the two quadrangles only. In order to minimize the calculation error of the cross-ratio due to undesirable marker pasting and to ensure a high decoding success rate, a large number of simulation and physical experiments are carried out to obtain the best cross-ratio categorization. As a result, the cross-ratio of the long side of CylinderTag is divided into four categories: 1.47, 1.54, 1.61, and 1.68. Users can adjust these values according to their needs. Unfortunately, the cross-ratio is of the nature of symmetry, which means that swapping two rectangles in the feature subregion yields the same result. It is worth noting that ``feature subregions'' under the same marker will always remain in the same direction, so there will only be two possible encodings of a marker in either direction. The symmetry of the cross-ratio can be dealt with by encoding the distribution of short and long edges
\begin{equation}
\delta = \begin{cases}0, & \frac{d_2}{d_3}>\frac{d_1+d_2}{d_3+d_4} \\ 1, & \frac{d_2}{d_3}<\frac{d_1+d_2}{d_3+d_4}\end{cases}
\end{equation}
where $\delta$ is the indicator of the distribution of short and long edges and $d_1 \cdots d_4$ are the distances of feature corners shown in Fig.~\ref{Fig:position}. In addition, the cross-ratios of the two long edges of the feature subregion are independent and the two cross-ratios participate in the encoding together, thus increasing the number of codes. The encoding of each feature subregion is formulated as follows
\begin{equation}
\begin{aligned}
& \{cr\} \rightarrow\{00,01,10,11\} \\
& C_{line}=(\delta<<2) \mid cr \\
& C_{feature}=(C_{line}^l<<1) \mid C_{line}^r
\end{aligned}
\end{equation}
where $C_{line}$ is the code of a long edge, each feature subregion corresponds to the two codes $C_{line}^l$, $C_{line}^r$, and the feature code $C_{feature}$ is obtained according to these two codes. However, the code count of individual ``feature subregions'' makes it difficult to support large-scale scenario applications. Therefore, similar to the idea of marker fields \cite{Bib:Herout2013CVPR,Bib:Zhu2022PAMI}, CylinderTag introduces ``feature fields'' that promise that $k$ consecutive encodings of feature subregions can only decode a unique marker ID, thus exponentially increasing the size of the marker dictionary.

CylinderTag offers an additional advantage of providing a larger number of features for localization as compared to traditional 4-corner quad markers. As described in \cite{Bib:Collins2014IJCV}, the increased number of features provides a better estimate of the object pose result for a variety of PnP algorithms.

\begin{figure*}[!t]
	\centering
    \includegraphics[width=.9\textwidth]{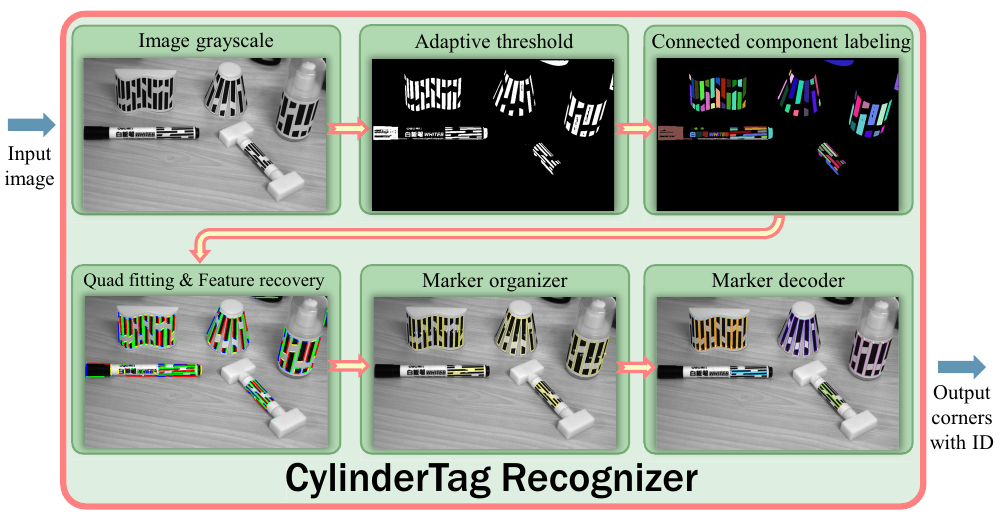}
	\caption{The pipeline of CylinderTag recognizer.}
    \label{Fig:pipeline}
\end{figure*}

\begin{algorithm}[!t]
 \caption{Heuristic Search Based Marker Generator}
 \label{Agr::Generation}
 \begin{algorithmic}[1]
  \REQUIRE Conflict library $\{\textit{C}\}$, marker size $n$, marker length $l$, feature length $f$
  \ENSURE Marker ID dictionary $\{\textit{M}\}$
  \WHILE{size($\{\textit{M}\}$) $< n$}
    \IF{size($M'$)$ == l$}
        \STATE $\{\textit{M}\} \leftarrow $\{\textit{M}\}$ + M'$
        \PRINT
    \ELSIF{$M' == \emptyset$}
        \STATE code $\leftarrow$ rand($64^f$)
        \WHILE{code $\not\in \{\textit{C}\}$}
            \STATE code $\leftarrow$ rand($64^f$)
        \ENDWHILE
        \STATE $\textit{M'} \leftarrow $ Convert code into octal
    \ELSIF{size($M'$)$ < n - 1$}
        \FOR{$i \in 0, \ldots, 64$}
            \IF{Legal($i$)}
                \STATE $b_{i} \leftarrow$ judgeConflict$(\textit{M'}, i)$
            \ENDIF
        \ENDFOR
        \STATE $\textit{M'} \leftarrow \textit{M'} + \mathop{\arg\max}\limits_{k} b_{k}$
    \ELSE
        \WHILE{Legal($c$) and CyclicConflict($\textit{M'},c$)}
            \STATE $c \leftarrow$ rand($64$)
        \ENDWHILE
        \STATE $\textit{M'} \leftarrow \textit{M'} + c$
    \ENDIF
  \ENDWHILE
 \end{algorithmic}
\end{algorithm}

\subsubsection{Generation Algorithm}
A ``feature field'' based marker generator can often be implemented using a search algorithm that takes an arbitrary feature as a node and iteratively generates a marker by judging the conflict between the marker and the current feature. However, traditional brute-force searching suffers from low efficiency and a tendency to fall into local backtracking for large coding numbers. We introduce the heuristic idea to optimize the search process. Specifically, at each step of updating markers, the number of potential solutions for each legal feature is counted, \ie, the number of legal update solutions that exist after selecting the current feature, and the feature with the highest number of potential solutions is selected for updating to maximize the feature utilization and success rate.

Marker generation is often a time-consuming operation. Many works have proposed various types of marker generation algorithms to obtain marker dictionaries in a fast and robust manner. Although marker generation only needs to be done offline, requiring less efficient generation algorithms, slow and inefficient generation algorithms still affect users. As mentioned in \cite{Bib:Garrido2014PR}, ArUco takes about 8, 20, and 90 minutes for dictionaries of size 10, 100, and 1000 respectively, while for AprilTag, it can take several days to generate a 36-bit marker. Our marker generator enables efficient mass generation of markers thanks to a large pool of codes and heuristic searching concepts. In our use, it takes only 20 seconds to generate over 1000 markers.

\subsection{Marker Recognizer}
Fig.~\ref{Fig:pipeline} shows the pipeline of the marker recognizer. To reduce the effects of noise or image blurring, the input grayscale image is downsampled to a certain scale depending on the original image size. Test results show that this procedure gives better recognition performance and a larger speed benefit compared to the original image.

\subsubsection{Adaptive Threshold}
For features consisting of high-contrast regions, adaptive threshold provides good segmentation of feature regions from the background by changing the input grayscale image into the binary image. Inspired by AprilTag 2, the input image is divided into tiles of $5\times5$ pixels to reduce computational cost compared with traditional methods. For each tile, the extrema are computed and stored. To avoid the influence of extreme pixels occurring in tile boundaries, the extrema pixel values $p_{\max}$ and $p_{\min}$ are defined as the extrema over $3\times3$ neighbor tiles. Thus, each pixel is given black or white in accordance with the threshold value defined as $( p_{\max} + p_{\min} ) / 2$.

\subsubsection{Connected-component Labeling}
After obtaining the binary image, the feature subregions are set to black, whereas the surrounding background is expected to become white. Therefore, each feature subregion should be divided into two isolated quads. Here, a fast and accurate connected components labeling algorithm named BBDT proposed by Grana \etal \cite{Bib:Grana2010TIP} is introduced. By establishing this algorithm, all the connected regions are congregated with different labels. The size of features is taken into account so that connected areas with a disproportionately large or small proportion of pixels are automatically filtered out.

\subsubsection{Quad Fitting}
\begin{figure}[!t]
\centering
\includegraphics[width=.35\textwidth]{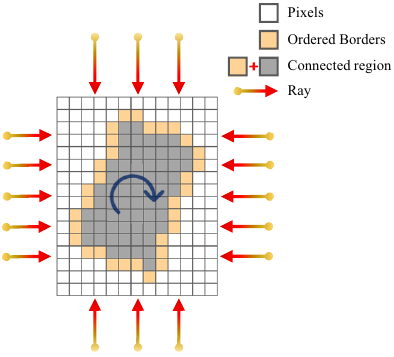}
\caption{\textbf{2D ray casting schematic.} Each grid represents a pixel, the yellow and gray grids are the connected regions acquired in the CCL algorithm, and the yellow grids are the acquired ordered border pixels.}
\label{Fig:raycasting}
\end{figure}
To recover the feature subregions, it is crucial to extract reasonable quads while excluding connected regions. However, traditional quad-fitting algorithms often suffer a high time complexity and are not suitable for use in CylinderTag, which is composed of a large number of quadrilaterals. To enhance the performance of the marker recognizer, we propose a fast quad-fitting mechanism called Extended Ramer-Douglas-Peucker (ERDP) algorithm. Ramer-Douglas-Peucker (RDP) algorithm is an iterative algorithm widely used in the GIS field for trajectory data compression. Its idea of iterative search is applied to the proposed quad-fitting algorithm. First, the border pixels of all connected regions are extracted by the 2D ray casting algorithm as shown in Fig.~\ref{Fig:raycasting}. Next, we rearrange the border pixels in connected order. A randomly selected border pixel is used as the starting point, and the pixels that are at the border are arranged in order using depth-first search (DFS).

\begin{algorithm}[!t]
 \caption{Extended Ramer-Douglas-Peucker Algorithm}
 \renewcommand{\algorithmicrequire}{\textbf{Input:}}
 \renewcommand{\algorithmicensure}{\textbf{Output:}}
 \renewcommand{\algorithmicprint}{\textbf{break}}
 \label{algorithm2}
 \begin{algorithmic}[1]
  \REQUIRE Ordered border pixel cluster $\{P\}$
  \ENSURE Four line functions $\{L\}$
  \STATE Calculate center $c$ of $\{P\}$
  \STATE $n \leftarrow$ the size of $\{P\}, s \leftarrow 0$
  \WHILE{size$\{L\}$ != 4}
   \WHILE{$cost > T_{cost}$}
    \STATE Calculate $cost \leftarrow \frac{\left\|\mathbf{p}_s+\mathbf{p}_{s+2}-2 \mathbf{p}_{s+1}\right\|_2}{\left\|\mathbf{p}_{s}-\mathbf{p}_{s+2}\right\|_2}$
    \IF{$cost > T_{cost}$}
     \STATE $n \leftarrow n+1$
    \ENDIF
   \ENDWHILE
   \WHILE{True}
       \STATE $e \leftarrow (s + n / 2) \% n$
       \STATE Calculate line $l \leftarrow$ Fit line($p_{s}, p_{e}$)
       \STATE Calculate distances $d_{i, i=s+1,...,e-1}$ to line $l$
       \IF{$\max d_{i} < T_{line}$}
        \STATE $\{P'\} \leftarrow$ Expand line($p_{s}, p_{e}$)
        \STATE $\{P\} \leftarrow \{P\} \backslash \{P'\}$
        \STATE $\{L\} \leftarrow \{L\} +$ Fit line($\{P'\}$)
        \PRINT
       \ELSE
        \STATE $e \leftarrow \mathop{\arg\max}\limits_{s} d_{s}$
       \ENDIF
       \ENDWHILE
  \ENDWHILE
 \end{algorithmic}
\end{algorithm}
As the algorithm shows, the border closest to the center is selected as the starting point of the ordered border, and the $cost=\frac{\left\|\mathbf{p}_s+\mathbf{p}_{s+2}-2 \mathbf{p}_{s+1}\right\|_2}{\left\|\mathbf{p}_{s}-\mathbf{p}_{s+2}\right\|_2}$ of the triplets connected to the current starting point is cyclically judged. The point that satisfies $cost \leq T_{cost}$ is considered the starting point. The point with the farthest distance from the ordinal number is selected as the endpoint, and the distance between the middle point and the line connecting these two points is obtained. If the maximum distance does not exceed the threshold $T_{line}$, it is considered a legal border, and the expansion module is entered to expand this border until both ends; if the maximum distance exceeds $T_{line}$, the border corresponding to the maximum distance is treated as the new endpoint, and the process is repeated until four legal borders are obtained.

Finally, for the candidate regions where the border pixel cluster is successfully recovered, we fit lines to the pixels of the four clusters to obtain four line functions. Thus the intersections of the lines can be calculated and considered as feature corners. Moreover, an indicator called Regional Area Coverage ($RAC=|S-Card(\{p_{i}\})|/S$, where $S$ refers to the area of the quad and $Card(\{p_{i}\})$ refers to the number of pixels belonging to this candidate region) is introduced, candidate regions with $RAC$ higher than the threshold $T_{RAC}$ are considered to deviate substantially from the quads and are thus rejected.

\subsubsection{Feature Recovery}
\begin{figure}[!t]
\centering
\includegraphics[width=.35\textwidth]{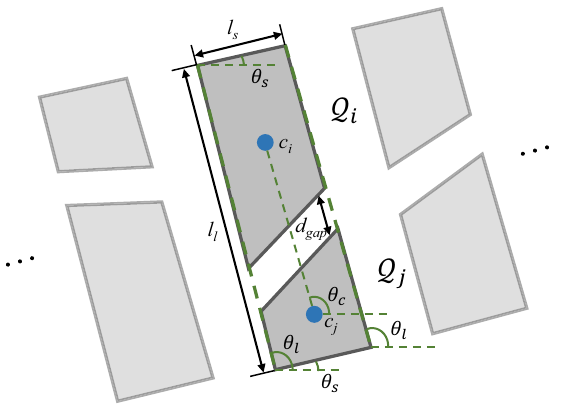}
\caption{Schematic diagram of the variables used in feature recovery.}
\label{Fig:featurerecovery}
\end{figure}

At this point, we have obtained all the candidate quad corners, and the next step is to select the quad pairs that belong to the same feature subregion. Geometrically there exist several conditions as follows to assist in filtering two quads $\mathcal{Q}_{i}$ and $\mathcal{Q}_{j}$ of the same feature subregion. First, the difference between the corresponding long side angles $\theta_{l}$ of the two quads should not exceed a threshold $T_{\theta}$, and both should be similar to the angle of the line connecting the centers of these two quads $\theta_{c}$. Second, since the short sides of the two quads are close to each other, the two short side angles $\theta_{s}$ should remain parallel. In addition, the difference between the long sides $l_{l}$ of two quads in a standard feature subregion and the distance between the quads $d_{gap}$, \ie, the width of the middle gap area, should be kept under a certain ratio $\alpha_{gap}$. Finally, the difference between the lengths of the two short edges $l_{s}$ and the lengths of the two long edges of the standard feature subregion should be kept under a certain ratio $\alpha_{len}$, and the difference between the lengths of the two short edges should be lower than the ratio $\alpha_{s}$. Fig.~\ref{Fig:featurerecovery} provides a detailed representation of the specific meanings of the aforementioned variables. The above criteria can be summarized as follows:
\begin{equation}
    \left\{\begin{aligned}
            \left|\Delta_{\theta_{l}}\right|  &\leq T_\theta\\
            \left|\Delta_{\theta_{s}}\right| &\leq T_\theta\\
            \left|\delta_{\theta_{i}}\right| &\leq T_\theta\\
            \left|\delta_{\theta_{j}}\right| &\leq T_\theta\\
            d_{gap} &\leq \alpha_{gap} * \sigma_{l}\\
            \sigma_{s} &\leq \alpha_{len} * \sigma_{l}\\
            \left|\Delta_{l_{s}}\right| &\leq \alpha_{s} * \min(l_{s_i}, l_{s_j})
        \end{aligned}
    \right.
\end{equation}
where $\delta_{\theta_{i}}=\theta_{l_i}-\theta_{c}$, $\delta_{\theta_{j}}=\theta_{l_j}-\theta_{c}$, and $\sigma_{l}=l_{i} + l_{j}$. Therefore, by filtering the above conditions, each quad can be enumerated to obtain a unique corresponding quad, thus forming a feature subregion as illustrated in Fig.~\ref{Fig:pipeline}.

\subsubsection{Edge Refinement}
To further improve the localization accuracy, a gradient-based double-weighted edge refinement method is implemented to obtain higher corner position accuracy. Due to the manifold assumption of a cylindrical surface, the quad of the feature subregion should lie approximately on a plane, which means the edges of the quad should remain straight. Since the inside of the quad of the feature subregion is uniformly black and the outside is white. For each corner, we uniformly sample its two adjacent edges and sample the image gradient along the normal edge to find the location of the largest image gradient in the current sampling point. However, poor placement when pasting the marker may cause the long edges of CylinderTag to be affected by curved surface deformations, making it difficult to maintain the straight-line assumption. A very intuitive idea is that edges close to corners are more likely to carry more accurate edge information, so the discrete point with the largest gradient is double-weighted according to the distance from the current corner and the magnitude of the gradient. The edge functions after refinement are obtained using PCA, and the refined feature corners are obtained based on the intersection of these edge functions.

\subsubsection{Marker Organization \& Feature Extraction}
Feature subregions belonging to the same marker should have close area size and angle, and the neighboring feature subregions should maintain the property of parallel long edges due to continuity. For two feature subregions $f_{i}$ and $f_{j}$, they are considered to belong to the same marker when the following properties are satisfied:
\begin{equation}
\left\{\begin{aligned}\label{Eqn:markerorganization}
  \left|\Delta_{\theta_{f}}\right| &\leq T_\theta \\
    \overrightarrow{l_i} \cdot \overrightarrow{c_i c_j} &\leq T_{ver} \\
    \overrightarrow{l_j} \cdot \overrightarrow{c_i c_j} &\leq T_{ver}
\end{aligned}\right.
\end{equation}
where $\theta_{f}$ refers to the long-edge angle of the feature subregion, $\overrightarrow{l_i}$ and $\overrightarrow{l_j}$ refer to the long-edge vectors of the two feature subregions, and $c_i$ and $c_j$ refer to the center of the two feature subregions.

Therefore, we introduce a marker organization method based on the Union-find algorithm to search for feature subregions that satisfy the above properties and cluster them. Once the markers are organized, we recover the cross-ratio of each feature subregion by extracting the edge lengths to obtain the corresponding ID. The corresponding subfigure in Fig.~\ref{Fig:pipeline} shows the results of marker organization and feature extraction.

\subsubsection{Marker Decoder}
For any marker, the center positions of the feature subregions are sorted and the encoding vector $\mathbf{c}^{*}$ is recovered according to the ratio of the center distance and the length of the short edge of the feature subregion. The encoding vector is matched with the marker ID dictionary $C$ for maximum coverage in both directions, and the dictionary position with the highest matching degree is the marker decoding result as shown in Fig.~\ref{Fig:pipeline}.

\subsection{3D Pose Recovery}
\begin{figure}[!t]
\centering
\includegraphics[width=.4\textwidth]{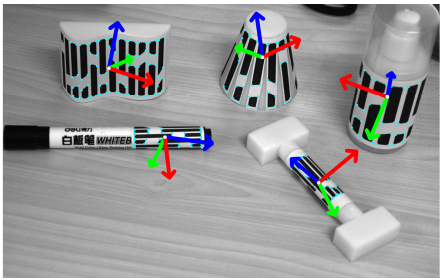}
\caption{Schematic diagram of curved object pose estimation.}
\label{Fig:pose}
\end{figure}

\subsubsection{Model Reconstruction}
CylinderTag is mainly oriented to the estimation of the pose of curved objects, so compared with the a priori knowledge that the traditional marker is located on a plane, this system often needs first to obtain the object model $\left\{\boldsymbol{P}^{W}_k \in \mathbb{R}^3 \mid k=1, \ldots, m\right\}$ corresponding to the current marker with $m$ feature corners and recover the object pose according to its model. This 3D reconstruction process can be accomplished using many classical SfM algorithms or software such as COLMAP. Here, we introduce an iterative 3D reconstruction pipeline \cite{Bib:Hu2023TASE}. First, a calibrated stereo vision system is used to take a surround shot of the object to be reconstructed, and $N$ pairs of stereo images $\left\{\left(I_{l i}, I_{r i}\right) \mid i=1, \ldots, N\right\}$ are obtained. Next, an iterative optimization algorithm is introduced to minimize the following reprojection error function
\begin{equation}
\boldsymbol{P}^{W *}=\underset{\boldsymbol{P}^W}{\arg \min } \sum_{i \in\left\{\left(I_{l i}, I_{r i}\right)\right\}}^{\boldsymbol{p}_{i k} \in C_{l i} \cup C_{r i}}\left\|\boldsymbol{p}_{i k}-\frac{1}{z_{i k}} \boldsymbol{K} \boldsymbol{T}_i \boldsymbol{P}_k^W\right\|_2^2
\end{equation}
where $\boldsymbol{p}_{i k}$ refers to the corners in $(I_{li}, I_{ri})$, $z_{ik}$ refers to the depth of $\boldsymbol{p}_{i k}$, and $\boldsymbol{K}$ is the intrinsic matrix.

\subsubsection{Pose Estimation}
After obtaining the 3D model $\{\boldsymbol{P}^{W}_k\}$ of the object, for any camera with known intrinsic parameters, we can obtain the matching relationship between the feature points in the captured image and the spatial location of the object model based on the decoding result of CylinderTag in the image, so as to optimize the following equations using the PnP algorithm and solve the poses $\boldsymbol{T}^*$ of the object in the camera coordinate.
\begin{equation}
\boldsymbol{T}^*=\underset{\boldsymbol{T}}{\operatorname{argmin}} \frac{1}{2} \sum_{i=1}^n\left\|\boldsymbol{p}_{\boldsymbol{i}}-\frac{1}{z_i} \boldsymbol{K} \boldsymbol{T} \boldsymbol{P}_i^W\right\|_2^2
\end{equation}

\section{Experimental Validation}
This section gives evaluations of CylinderTag versus traditional marker systems and shows several application scenarios for CylinderTag. The cameras used are HikVision MV-CA023-10GM industrial cameras (monochrome, 1920$\times$1200 pixels). The computer is equipped with Intel I7-11700K (2.50~GHz). Also, the hyperparameters for the recognizer of CylinderTag are set as follows: $T_{cost}=\text{1.05}$, $T_{line}=\text{1.8}$, $T_{RAC}=\text{0.3}$, $T_{\theta}=\text{5}^{\circ}$, $\alpha_{gap}=\text{0.067}$, $\alpha_{len}=\text{15}$, $\alpha_{s}=\text{0.33}$, and $T_{ver}=\text{0.5}$.

\subsection{Detection Accuracy}
\begin{table}[!t]
\centering
\caption{False Positive of Different Marker Systems in Indoor Scene Recognition Dataset}
\begin{tabular}{c|c|c}
\hline
Marker               & False number & FP           \\
\hline
AprilTag (16h5)      & 6631         & 42.452\%  \\
AprilTag (25h9)      & 79           & 0.506\%   \\
AprilTag (36h11)     & \textbf{0}   & \textbf{0\%} \\
ArUco\_ORIGINAL\_5$\times$5 & 704          & 4.507\%   \\
ArUco\_4$\times$4\_250      & 699          & 4.475\%   \\
ArUco\_5$\times$5\_100      & 5            & 0.032\%   \\
ArUco\_6$\times$6\_250      & 1            & 0.006\%   \\
RUNE-Tag             & \textbf{0}   & \textbf{0\%} \\
TopoTag (3$\times$3)        & 2            & 0.013\%   \\
TopoTag (4$\times$4)        & 2            & 0.013\%   \\
CylinderTag (12c2f)  & \textbf{0}   & \textbf{0\%} \\
CylinderTag (15c3f)  & \textbf{0}   & \textbf{0\%} \\
\hline
\end{tabular}
\label{Tab:falsepositive}
\end{table}

Detection accuracy is the most important metric to evaluate the effectiveness of a marker system in use. In this experiment, we compared the detection accuracy of CylinderTag with four traditional planar marker systems (AprilTag 3 \cite{Bib:Maximilian2019IROS} , ArUco \cite{Bib:Garrido2014PR}, TopoTag \cite{Bib:Yu2020TVCG} , and RUNETag \cite{Bib:Bergamasco2016TPAMI}) under different viewpoints. True positive ($TP$) is defined as a correct detection of the marker, which contains both a successful location of the marker and a correct decoding of the ID. False positive ($FP$) is defined as a detection result returned by the detection algorithm that does not correctly identify the location or ID. False negative ($FN$) is defined as the presence of a marker at a location that is not correctly identified with any marker. Furthermore, precision is defined as $\frac{TP}{TP+FP}$ while recall is defined as $\frac{TP}{TP+FN}$.

First, the Indoor Scene Recognition dataset \cite{Bib:Quattoni2009CVPR} with 15,620 images from 67 categories to represent unmarked scenes is introduced to measure the $FP$ of CylinderTag versus traditional marker systems. Table~\ref{Tab:falsepositive} shows that both Apriltag and ArUco rely heavily on their error correction capabilities for false rejection. Therefore, the $FP$ decreases significantly as the Hamming distance increases. TopoTag detects only 2 false markers due to its tight internal topological structure arrangement. No false marker is detected by RUNETag, which is attributed to its extremely strong error correction capability. CylinderTag does not detect any false positives either due to the unique design of the features and the encoding concept of ``feature field''.

\begin{figure}[!t]
\centering
\includegraphics[width=.4\textwidth]{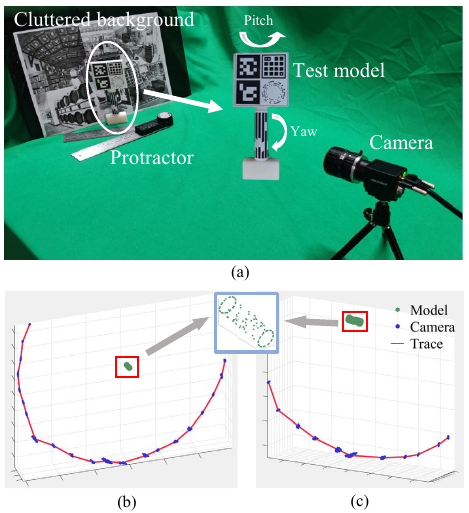}
\caption{\textbf{Experiment setup of detection accuracy.} (a) Real-world image acquisition process. (b) Camera pose and movement track for yaw-view experiments. (c) Camera pose and movement track for pitch-view experiments.}
\label{Fig:exp1setup}
\end{figure}

Further, a 3D printed model attached with traditional markers and CylinderTag is used to test the detection accuracy under cluttered backgrounds with different viewpoints. Here, we tested the precision and recall of each marker system under different viewing angles corresponding to the two axes of pitch and yaw with a fixed camera, and Fig.~\ref{Fig:exp1setup} plots the position and movement track of the camera relative to the model during the experiment. The specific test procedure is: after fixing the model to a fixed viewing angle, the camera is moved to the vertical direction in the visual range while acquiring around $200$ frames, and the detection algorithm is used to calculate the precision and recall of each marker system.

\begin{figure}[!t]
\centering
\includegraphics[width=.4\textwidth]{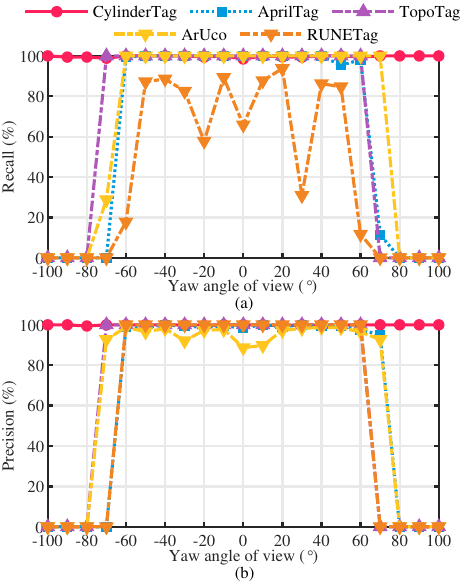}
\caption{Precision and recall of different marker systems at different yaw angles of view.}
\label{Fig:detectionyaw}
\end{figure}

\begin{figure}[!t]
\centering
\includegraphics[width=.4\textwidth]{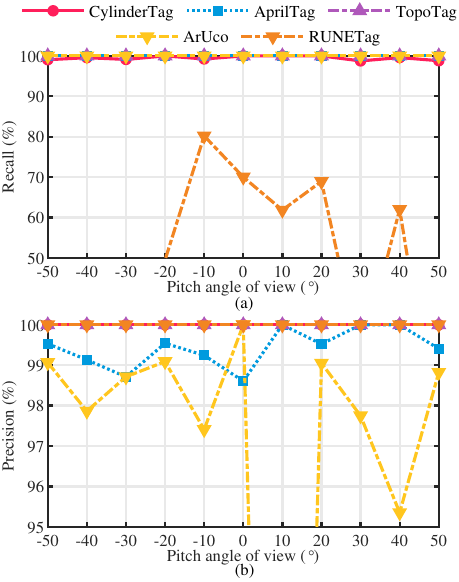}
\caption{Precision and recall of different marker systems at different pitch angles of view.}
\label{Fig:detectionpitch}
\end{figure}

For cylindrical-like objects, a similar viewing area tends to be maintained on the yaw axis. For planar markers, one of the biggest problems is that a larger viewing angle can lead to a significant reduction in the visible area, thus reducing the detection performance, as Fig.~\ref{Fig:detectionyaw} demonstrates. For quadrilateral markers such as AprilTag, ArUco, and TopoTag, the detection is not stable when the yaw axis angle is greater than $60^{\circ}$; for RUNETag, the precision is highly unstable because its features are made up of dots, which are difficult to be detected at small scales; CylinderTag, as a marker that tightly fits on the curved surface, can maintain a stable detection performance on the yaw axis, which reflects its detection advantages under different viewpoints.

In addition, experiments on the detection performance of the markers at different pitch angles are required. For planar markers, the yaw angle and pitch angle can be regarded as similar and can be reflected in Fig.~\ref{Fig:detectionyaw} and Fig.~\ref{Fig:detectionpitch}. For CylinderTag, due to the projection invariance of the quadrilateral feature, its stable detection performance can still be maintained under a certain pitch angle, which verifies its high detection rate under different viewing angles.

The results similar to Table~\ref{Tab:falsepositive} are also reflected in the precision results of various markers in this experiment. The precision of CylinderTag, TopoTag, and RUNETag all remain close to $100\%$. In contrast, AprilTag and ArUco both had some amount of false positives.

\subsection{Detection Speed \& Dictionary Size}
Detection speed and dictionary size are also two important metrics to measure the performance of the marker system. Table~\ref{Tab:speedandsize} lists the detection speed and dictionary size of different markers. CylinderTag, AprilTag, and ArUco all maintain $\sim\text{30}$ FPS real-time detection capability, while TopoTag maintains $\sim\text{8}$ FPS owing to the segmentation part; RUNETag achieves only $\sim\text{1.5}$ FPS due to the extensive and time-consuming ellipse fitting operation.

In addition, the application of large scenes poses a challenge to the dictionary size of the marker system. In order to compromise sufficient error correction capability and recognition performance, the dictionary size of AprilTag and ArUco cannot exceed one thousand; RUNETag can support the generation of 17000 markers due to its heptad expansion; TopoTag achieves direct mapping from binary to ID due to its topological structure, thus obtaining sufficient dictionary size as well; CylinderTag uses cross-ratio as the encoding method, and introduces the idea of ``feature field'' to expand the encoding, thus providing a large enough encoding amount. In summary, compared with the problems of traditional marker systems, CylinderTag can support real-time applications in large scenarios.
\begin{table}[]
\centering
\caption{Comparison of Detection Speed and Dictionary Size}
\begin{tabular}{c|c|c|c}
\hline
Marker                       & \begin{tabular}[c]{@{}c@{}}Detection\\ speed (ms)\end{tabular} & Type     & Dictionary size \\ \hline
\multirow{3}{*}{CylinderTag} & \multirow{3}{*}{35.21}                                         & 12c2f    & 41              \\ \cline{3-4}
                             &                                                                & 15c3f    & 1092            \\ \cline{3-4}
                             &                                                                & 18c4f    & 29098           \\ \hline
\multirow{3}{*}{AprilTag}    & \multirow{3}{*}{33.49}                                         & 16h5     & 30              \\ \cline{3-4}
                             &                                                                & 25h9     & 35              \\ \cline{3-4}
                             &                                                                & 36h11    & 587             \\ \hline
\multirow{2}{*}{TopoTag}     & \multirow{2}{*}{121.40}                                        & 3X3      & 128             \\ \cline{3-4}
                             &                                                                & 4X4      & 16384           \\ \hline
RUNE-Tag                     & 648.82                                                         & RUNE-129 & 17000           \\ \hline
\multirow{3}{*}{ArUco}       & \multirow{3}{*}{32.03}                                         & 16h3     & 250             \\ \cline{3-4}
                             &                                                                & 25h7     & 100             \\ \cline{3-4}
                             &                                                                & 36h12    & 250             \\ \hline
\end{tabular}
\label{Tab:speedandsize}
\end{table}

\subsection{Localization Jitter Evaluation}
The localization accuracy of CylinderTag in real-world scenarios was further evaluated by introducing jitter error as a metric to evaluate the marker system. In fields such as AR and VR, the violent jitter of the marker can lead to an unpleasant ``shaking'' effect on 3D objects that are rendered on top of the marker. To measure the localization accuracy of the marker, a cylindrical-shaped 3D model was designed to test the jitter error under different viewing angles and distances. Here, with a focus on detection success rates on the curved surface, we designed a ring of $8\times5$ AprilTag as a comparison model.

\begin{figure}[!t]
\centering
\includegraphics[width=.42\textwidth]{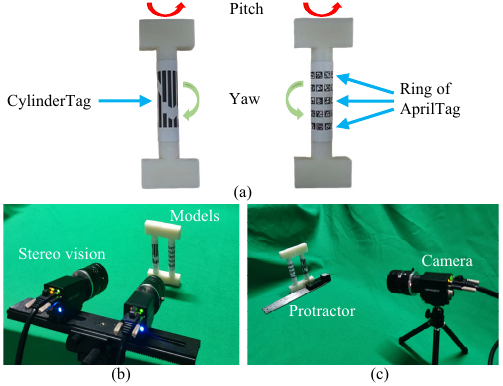}
\caption{\textbf{Experiment setup of localization accuracy.} (a) Schematic diagram of the 3D printed model. (b) 3D reconstruction process. (c) Real-world experimental acquisition process.}
\label{Fig:exp2setup}
\end{figure}

\begin{table}[]
\caption{Comparison of Input Images and Reprojection Errors in 3D Reconstruction}
\centering
\begin{tabular}{c|cc|c}
\hline
\multirow{2}{*}{Evaluation metrics} & \multicolumn{2}{c|}{CylinderTag}               & \multirow{2}{*}{AprilTag} \\ \cline{2-3}
                                    & \multicolumn{1}{c|}{w/ slope} & w/o slope      &                           \\ \hline
Input images                        & \multicolumn{2}{c|}{\textbf{20}}                        & 57                        \\ \hline
Triangulation RPE                   & \multicolumn{1}{c|}{0.236}    & \textbf{0.217} & 0.245                     \\ \hline
Left-view BA RPE                    & \multicolumn{1}{c|}{0.403}    & \textbf{0.121} & 0.122                     \\ \hline
Right-view BA RPE                   & \multicolumn{1}{c|}{0.288}    & \textbf{0.092} & 0.106                     \\ \hline
\end{tabular}
\label{Tab:reconstruction}
\end{table}

\subsubsection{Reconstruction}
First, we performed 3D reconstructions described in Section 3.3 of the two marker systems and compared the reprojection errors (RPE) of these two reconstructed models. As demonstrated in Table~\ref{Tab:reconstruction}, the reprojection error of the reconstruction model using CylinderTag is larger than that of AprilTag. Nevertheless, in subsequent experiments, we utilized the model to verify that its localization accuracy is significantly superior to that of AprilTag. This finding indicates that reprojection error may not be an accurate indicator of localization accuracy. Upon further examination of the CylinderTag model, we discovered that the reprojection error of corners corresponding to sloping edges was notably high. However, after eliminating these corners, the reprojection errors of the reconstructed model were all lower than those of AprilTag. This provides evidence that in the same conditions, the reconstruction accuracy of CylinderTag is superior. In addition, the compact structure of the CylinderTag allows for a convenient 3D reconstruction process which requires significantly fewer input images than required by the AprilTag.

\subsubsection{Under Different Viewing Yaw Angles}
\begin{figure}[!t]
\centering
\includegraphics[width=.4\textwidth]{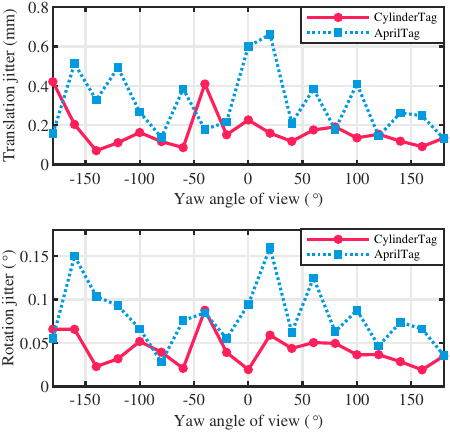}
\caption{Rotation and translation jitter errors at different yaw angles of view.}
\label{Fig:localizationyaw}
\end{figure}

\begin{figure}[!t]
\centering
\includegraphics[width=.4\textwidth]{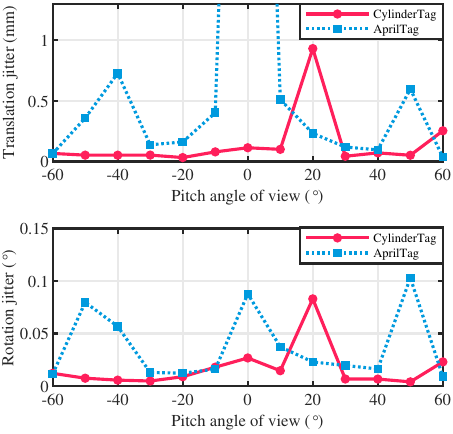}
\caption{Rotation and translation jitter errors at different pitch angles of view.}
\label{Fig:localizationpitch}
\end{figure}
The standard deviation (STD) metric was employed to evaluate the translation and rotation jitter for each yaw angle. Fig.~\ref{Fig:localizationyaw} depicts that CylinderTag exhibits significantly lower jitter errors than AprilTag. Furthermore, two-sample Kolmogorov-Smirnov tests indicate that CylinderTag surpasses AprilTag significantly in both rotation jitter ($p=\text{0.006}$) and translation jitter ($p=\text{0.000}$).

\subsubsection{Under Different Viewing Pitch Angles}
When dealing with markers on curved surfaces, the pitch angle and yaw angle cannot be considered interchangeable. Therefore, the assessment of translation and rotation jitter errors for the pitch angle was also carried out using the same evaluation process as that of the yaw angle. As shown in Fig.~\ref{Fig:localizationpitch}, CylinderTag displays substantially lower jitter errors than AprilTag. Additionally, two-sample Kolmogorov-Smirnov tests reveal that CylinderTag outperforms AprilTag significantly in both rotation jitter ($p=\text{0.028}$) and translation jitter ($p=\text{0.008}$).

\begin{figure}[!t]
\centering
\includegraphics[width=.4\textwidth]{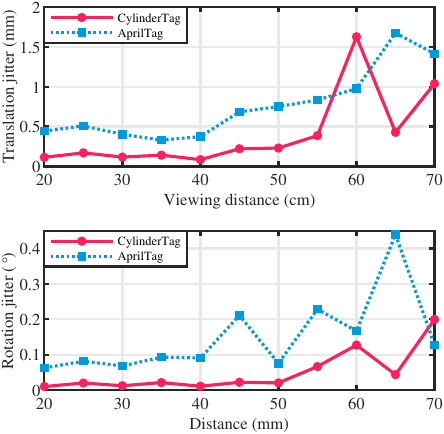}
\caption{Rotation and translation jitter errors at different viewing distances.}
\label{Fig:localizationdist}
\end{figure}

\subsubsection{Under Different Distances}
Different distances also have a significant impact on localization accuracy, and therefore, we assessed the jitter error for these two models at various distances. As depicted in Fig.~\ref{Fig:localizationdist}, CylinderTag displays significantly lower jitter errors than AprilTag. Moreover, two-sample Kolmogorov-Smirnov tests indicate that CylinderTag outperforms AprilTag significantly in both rotation jitter ($p=\text{0.002}$) and translation jitter ($p=\text{0.012}$).

\subsection{Pose Estimation Accuracy}
\begin{figure}[!t]
    \centering
    \subfloat[]{{\includegraphics[width=.42\textwidth]{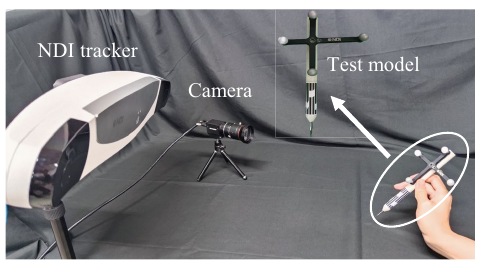}}\label{Fig:PoseExpa}}
    \\
    \vspace{-8pt}
    \subfloat[]{{\includegraphics[width=.15\textwidth]{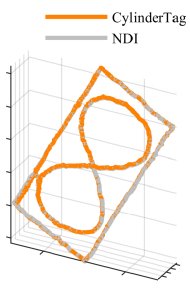}}\label{Fig:PoseExpb}}
    \subfloat[]{{\includegraphics[width=.29\textwidth]{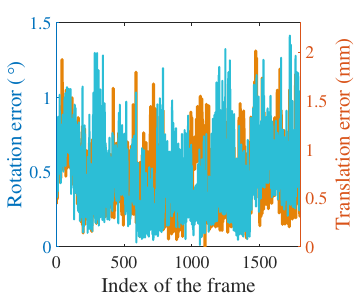}}\label{Fig:PoseExpc}}
    \caption{\textbf{Experiment setup and results of pose estimation accuracy.} (a) Setup diagram for the pose estimation experiment. (b) Test model tip trajectories measured with CylinderTag versus ground truth trajectories. (c) Translation and rotation errors for estimated poses.}
    \label{Fig:PoseEstExp}
\end{figure}

\begin{figure}[!t]
\centering
\includegraphics[width=.4\textwidth]{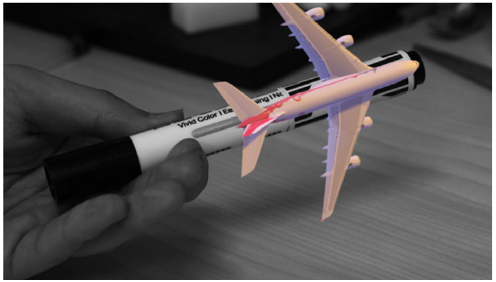}
\caption{\textbf{AR applications.} Acquire the pose of the pen by detecting the CylinderTag and display a model of the airplane in real time.}
\label{Fig:AR}
\end{figure}

For a more thorough evaluation of the pose estimation capability of CylinderTag, it is imperative to assess the accuracy of pose tracking for the CylinderTag attached to an object during motion. Here, similar to the methodology described in \cite{Bib:Zhang2017IJCARS}, we employed the NDI tracker (Polaris Vega tracker, Northern Digital, Inc.) to acquire ground truth pose information. The experiment setup for the pose estimation accuracy is shown in Fig.~\ref{Fig:PoseExpa}. First, a 3D-printed cylindrical model with a tip mounted on its end was designed for evaluation. Next, the transformation relationship of the NDI tracker w.r.t the camera was obtained by tip registration and hand-eye calibration. During the experiment, the tip pose detected by the monocular camera through CylinderTag and the tip pose measured by the NDI tracker were recorded simultaneously. 2000 consecutive frames were recorded and the translation and rotation errors of the estimated pose by CylinderTag with those obtained by the NDI were calculated.

Fig.~\ref{Fig:PoseExpb} shows the plot of the test model tip trajectory obtained via CylinderTag against the reference trajectory obtained by the NDI tracker, and Fig.~\ref{Fig:PoseExpc} shows the errors in translation and rotation of the corresponding estimated pose for each frame. The average pose estimation error was 0.69 mm with a standard deviation of 0.38 mm for the translation and $\text{0.46}^{\circ}$ with a standard deviation of $\text{0.20}^{\circ}$ for the rotation. Notably, the pose estimation error of CylinderTag is significantly smaller than the results reported in \cite{Bib:Zhang2017IJCARS}, where the best translation error was $\text{1.43}\pm\text{1.09}$ mm and the best rotation error was $\text{0.55}\pm\text{0.38}^{\circ}$.

\subsection{Applications}
\subsubsection{Augmented Reality}
Today's visual marker-based AR applications are still stuck on a flat surface, which greatly limits the interaction capability and application scenarios of AR. As shown in Fig.~\ref{Fig:AR}, CylinderTag provides feasibility for the implementation of the AR domain on cylindrical-like curved objects. By attaching CylinderTag to its surface, it can be photographed with a calibrated camera or cell phone camera to obtain its positional information, thus enabling more advanced augmented reality applications.

\subsubsection{Digital Pen}
\begin{figure}[!t]
\centering
\includegraphics[width=.4\textwidth]{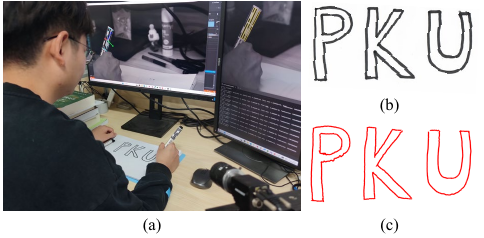}
\caption{\textbf{Digital pen applications.} (a) The digital pen turns a flat surface into a digital drawing surface. (b) Real handwriting obtained by the scanner. (c) Handwriting generated by the digital pen.}
\label{Fig:DP}
\end{figure}
The use of vision-based digital pens has significantly advanced the fields of VR/AR. These pens are equipped with visual marker modules that allow for precise and accurate 6-DoF tracking. As a result, they are widely used in applications such as VR gaming, 2D drawing, and 3D modeling. However, today's digital pens often require customization or additional physical modules \cite{Bib:Wu2017UIST}, which is inconvenient for users to use. With CylinderTag, users can quickly transform their pen into a digital pen that can track their posture in real-time, thus ``digitizing any pen.'' Fig.~\ref{Fig:DP} illustrates the process of transforming an ordinary pen into a digital pen with end-tracking capabilities.

\subsubsection{Surgical Robot Navigation}
\begin{figure}[!t]
\centering
\includegraphics[width=.4\textwidth]{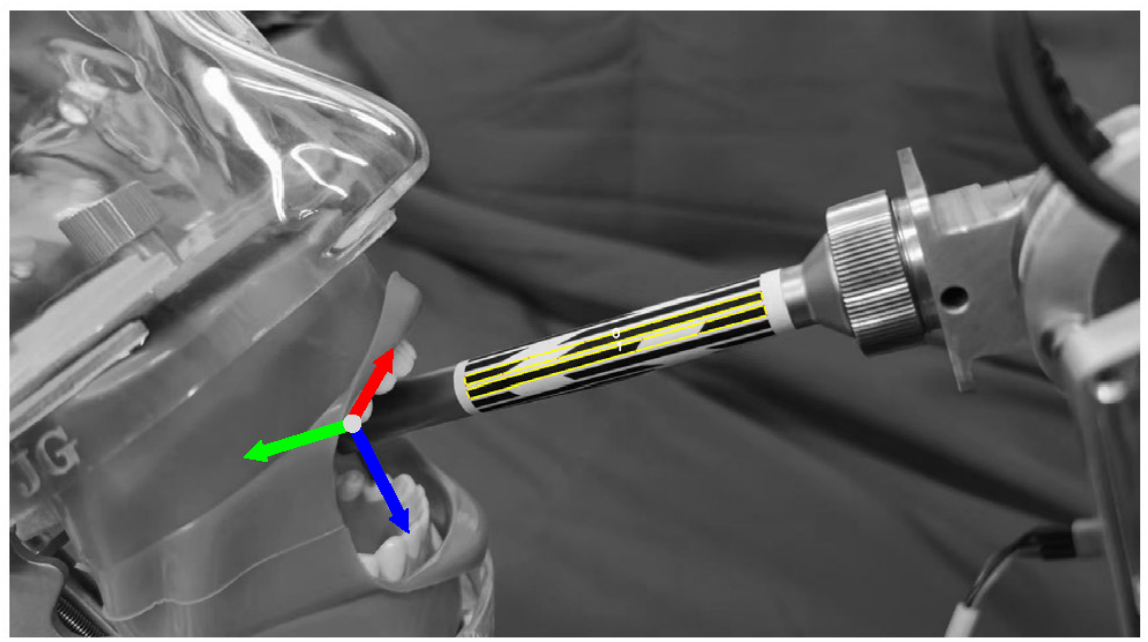}
\caption{\textbf{Surgical robot navigation applications.} Attaching CylinderTag tightly to the end of a surgical robot or instrument enables compact visual navigation.}
\label{Fig:Surgical}
\end{figure}
Visual navigation is an essential technology for achieving automated surgical robots \cite{Bib:Fan2017TBME,Bib:Zhang2017IJCARS}. Traditional visual navigation system is mainly based on an external infrared ball joint, which is unsuitable for surgeries with limited operating space, such as oral and pharyngeal surgeries, as well as minimally invasive surgeries due to their large space occupation. CylinderTag can be tightly attached to the cylindrical end-effector of the surgical robot, enabling high-precision visual navigation without occupying additional space shown in Fig.~\ref{Fig:Surgical}.

\subsection{Shortcomings \& Extensions}
\begin{figure}[!t]
\centering
\includegraphics[width=.4\textwidth]{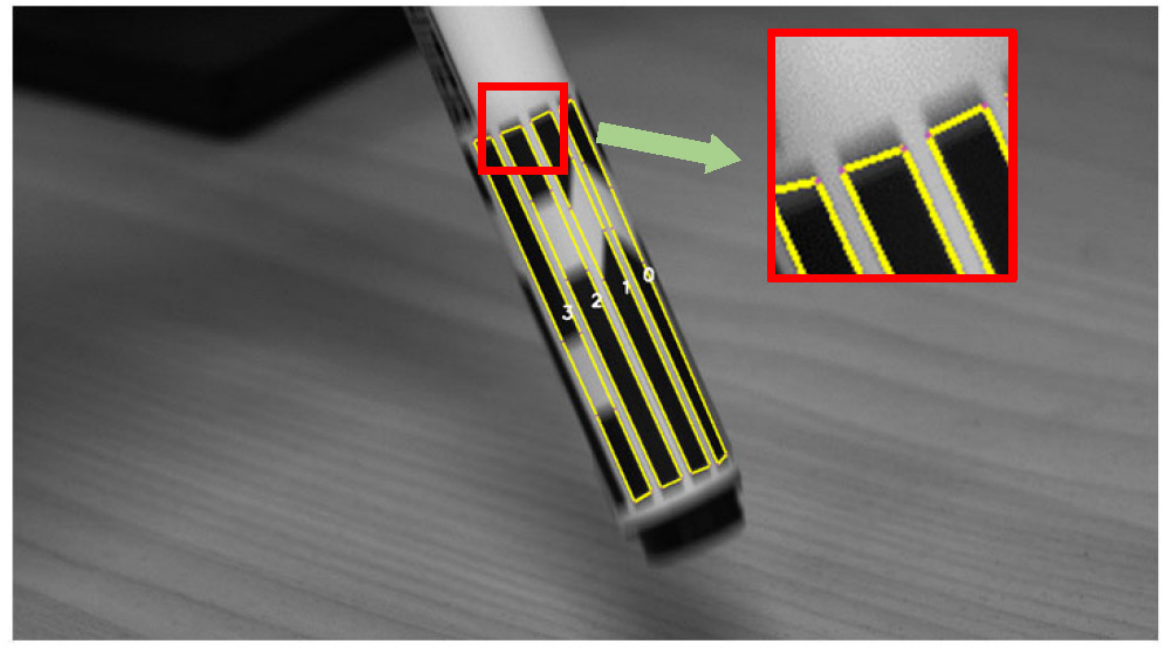}
\caption{\textbf{A failure case.} A CylinderTag encountering violent motion blur may cause the marker to be decoded incorrectly.}
\label{Fig:motionblur}
\end{figure}

\begin{figure}[!t]
\centering
\includegraphics[width=.4\textwidth]{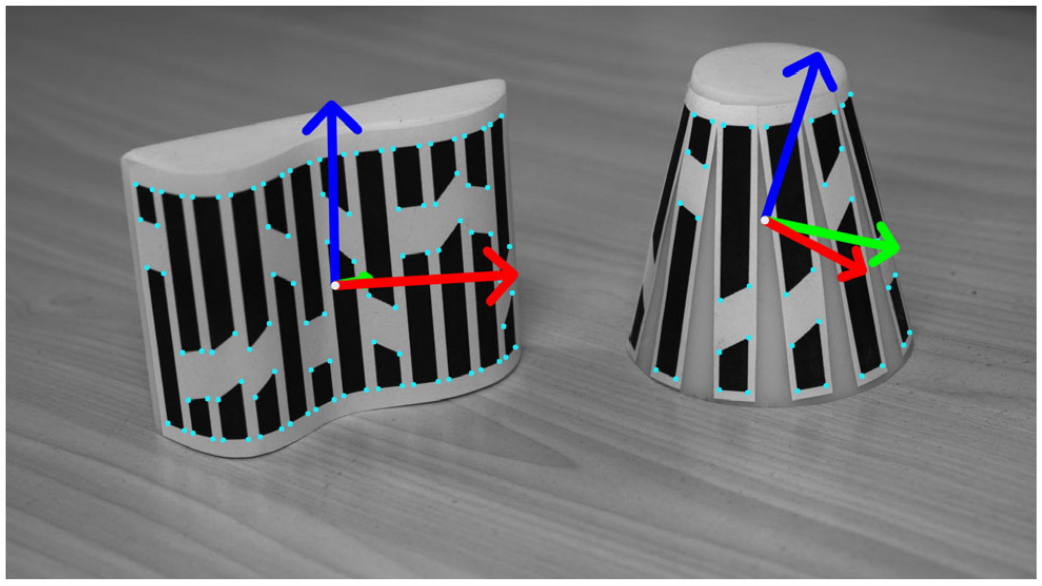}
\caption{\textbf{An extension case.} CylinderTag can be used to estimate poses for developable surfaces such as conical surfaces and some profiled surfaces.}
\label{Fig:extension}
\end{figure}
CylinderTag relies on the cross-ratio corresponding to the intersection of sub-pixel edges for encoding. However, when there is significant motion blur, as shown in Fig.~\ref{Fig:motionblur}, the edge equation extracted by the recognizer may be biased, resulting in incorrect decoding of the cross-ratio and ultimately leading to the wrong marker ID being recovered.

As CylinderTag is primarily crafted for the mentioned developable surfaces, with cylindrical surfaces being the most prevalent, the marker can also be extended to various other surface types, including domed surfaces, conical surfaces, and some profiled surfaces as shown in Fig.~\ref{Fig:extension}, which significantly broadens the application scope of CylinderTag.

\section{Conclusion}
High-precision localization of curved objects offers a promising future for AR, VR, and robotics. In this paper, we propose CylinderTag, a visual marker for cylindrical-like objects, which is based on the manifold assumption and encoded using cross-ratio, and the concept of feature field is introduced to expand the encoding capacity. In addition, a marker recognizer for real-time detection and recognition is proposed, and 3D reconstruction of curved objects is realized. Various experiments, including detection rate, detection speed, dictionary size, localization jitter evaluation, and pose estimation accuracy, have demonstrated the outstanding performance of CylinderTag. It is currently the most advanced visual marker that can be applied to curved objects with developable surfaces.

In the future, we will focus on enhancing the versatility of CylinderTag by developing a more robust recognizer to identify markers on various curved surfaces. We believe that CylinderTag can serve as a source of inspiration for the development of tight visual markers for arbitrary objects.

\vspace{-1cm}
\begin{IEEEbiography}[{\includegraphics[width=1in,height=1.25in,clip,keepaspectratio]{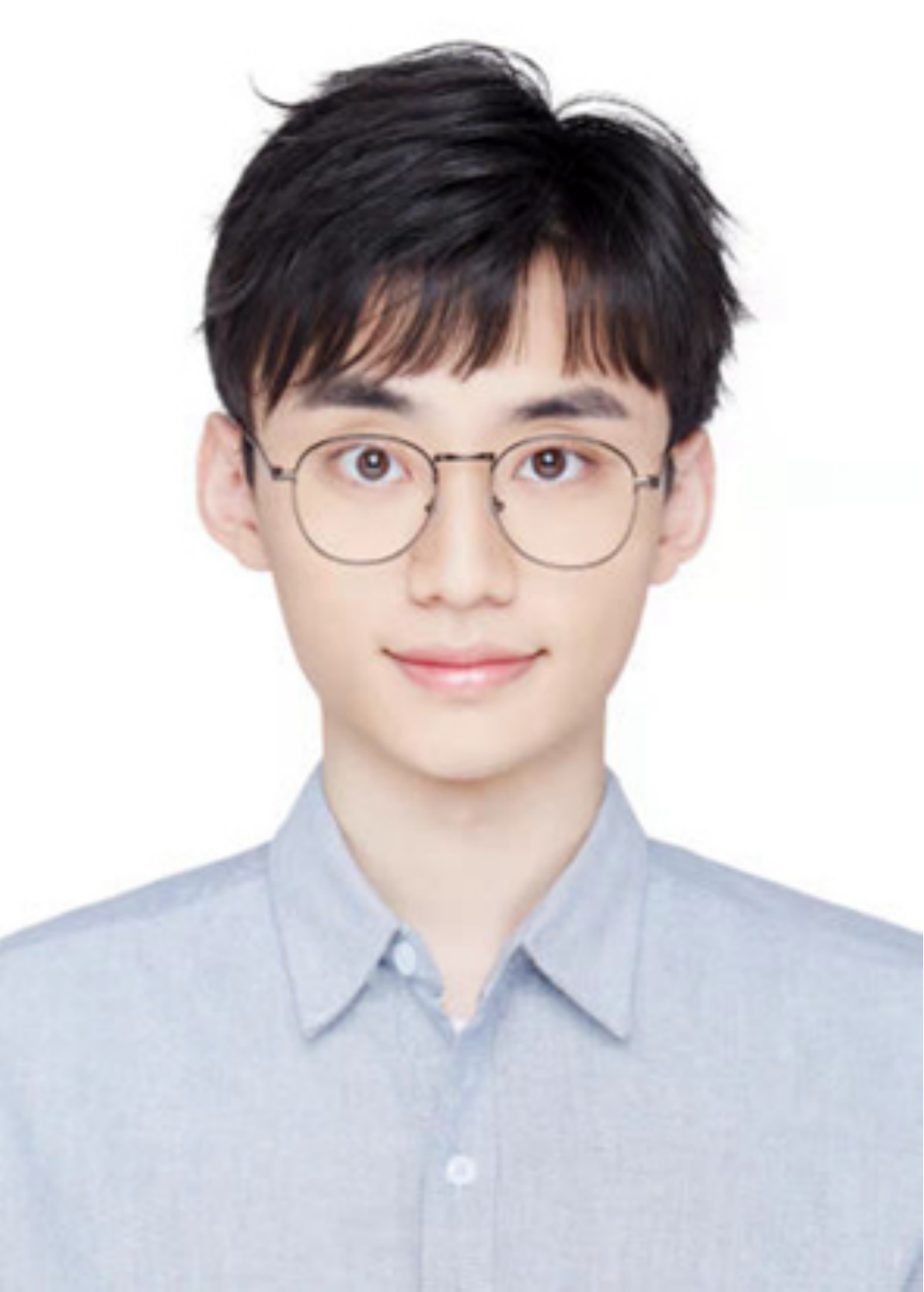}}]{Shaoan Wang}
received the B.E. degree in mechanical engineering from the School of Mechatronical Engineering, Beijing Institute of Technology, Beijing, China, in 2021. He is currently pursuing the Ph.D. degree in general mechanics and foundation of mechanics with the College of Engineering, Peking University, Beijing, China. His current research interests include robot vision and visual localization.
\end{IEEEbiography}

\vspace{-1cm}
\begin{IEEEbiography}[{\includegraphics[width=1in,height=1.25in,clip,keepaspectratio]{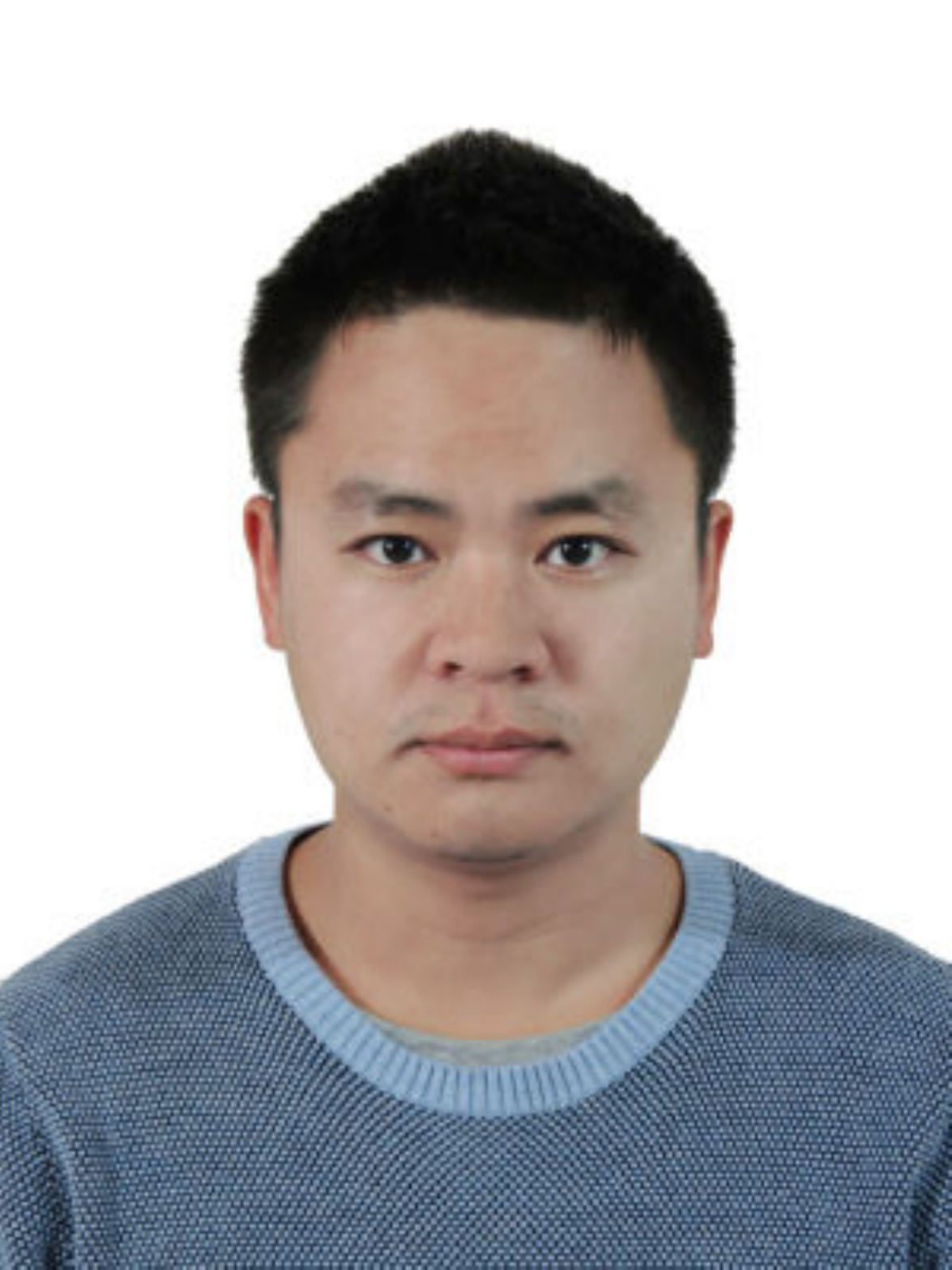}}]{Mingzhu Zhu}
received the B.E, M.E, and D.E. degrees in the School of Mechanical Engineering and Automation, Fuzhou University, Fuzhou, China, in 2012, 2015, and 2019, respectively. From 2019 to 2021, he was a Postdoctoral Research Fellow with BIC-ESAT, College of Engineering, Peking University, Beijing, China. In 2021, he joined Fuzhou University, as an Associate Scientist. His current research interests include computer vision and image processing.
\end{IEEEbiography}

\vspace{-1cm}
\begin{IEEEbiography}[{\includegraphics[width=1in,height=1.25in,clip,keepaspectratio]{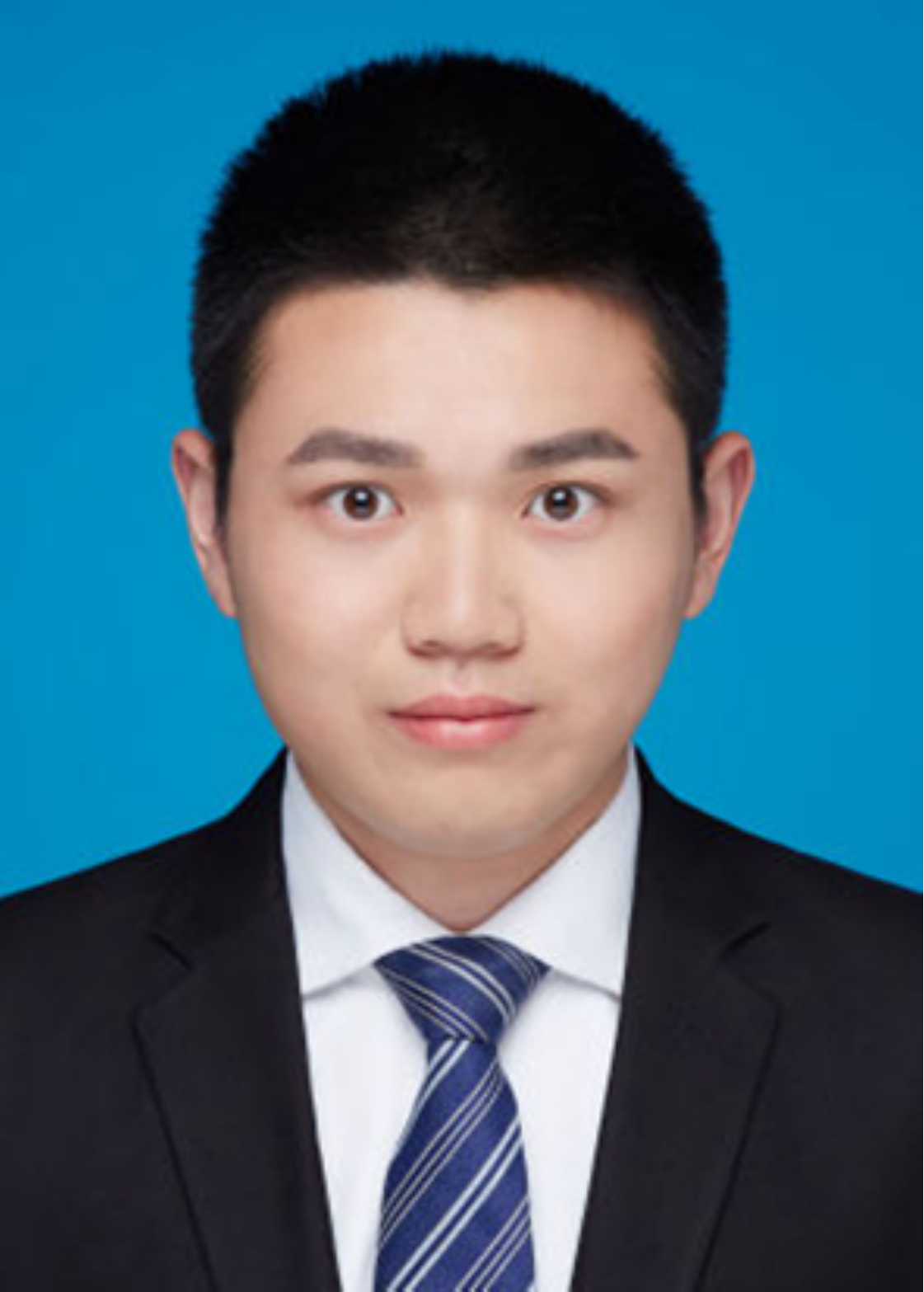}}]{Yaoqing Hu}
received the B.E. degree from University of Science and Technology Beijing, China, in 2018, and the M.E. degree from University of Science and Technology Beijing, China, in 2021. He is currently pursuing the Ph.D. degree in general mechanics and foundation of mechanics with the College of Engineering, Peking University, Beijing, China. His research interests include computer vision and robotics.
\end{IEEEbiography}

\vspace{-1cm}
\begin{IEEEbiography}[{\includegraphics[width=1in,height=1.25in,clip,keepaspectratio]{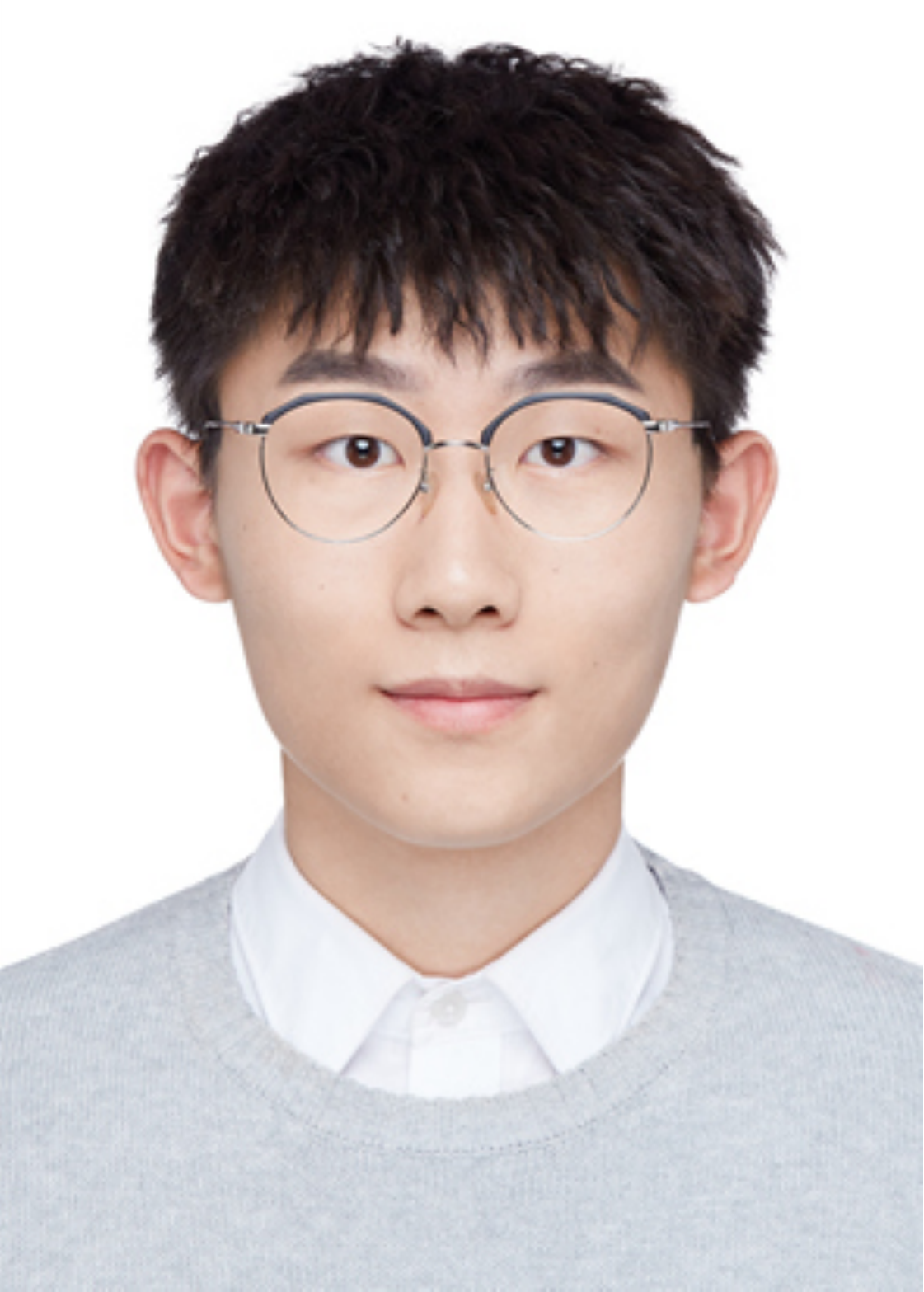}}]{Dongyue Li}
received the B.E. degree from the School of Aerospace, Beijing Institute of Technology, Beijing, China, in 2020. He is currently pursuing the Ph.D. degree in general mechanics and foundation of mechanics with the College of Engineering, Peking University, Beijing, China. His current research interests include robot vision and surgical robot.
\end{IEEEbiography}

\vspace{-1cm}
\begin{IEEEbiography}[{\includegraphics[width=1in,height=1.25in,clip,keepaspectratio]{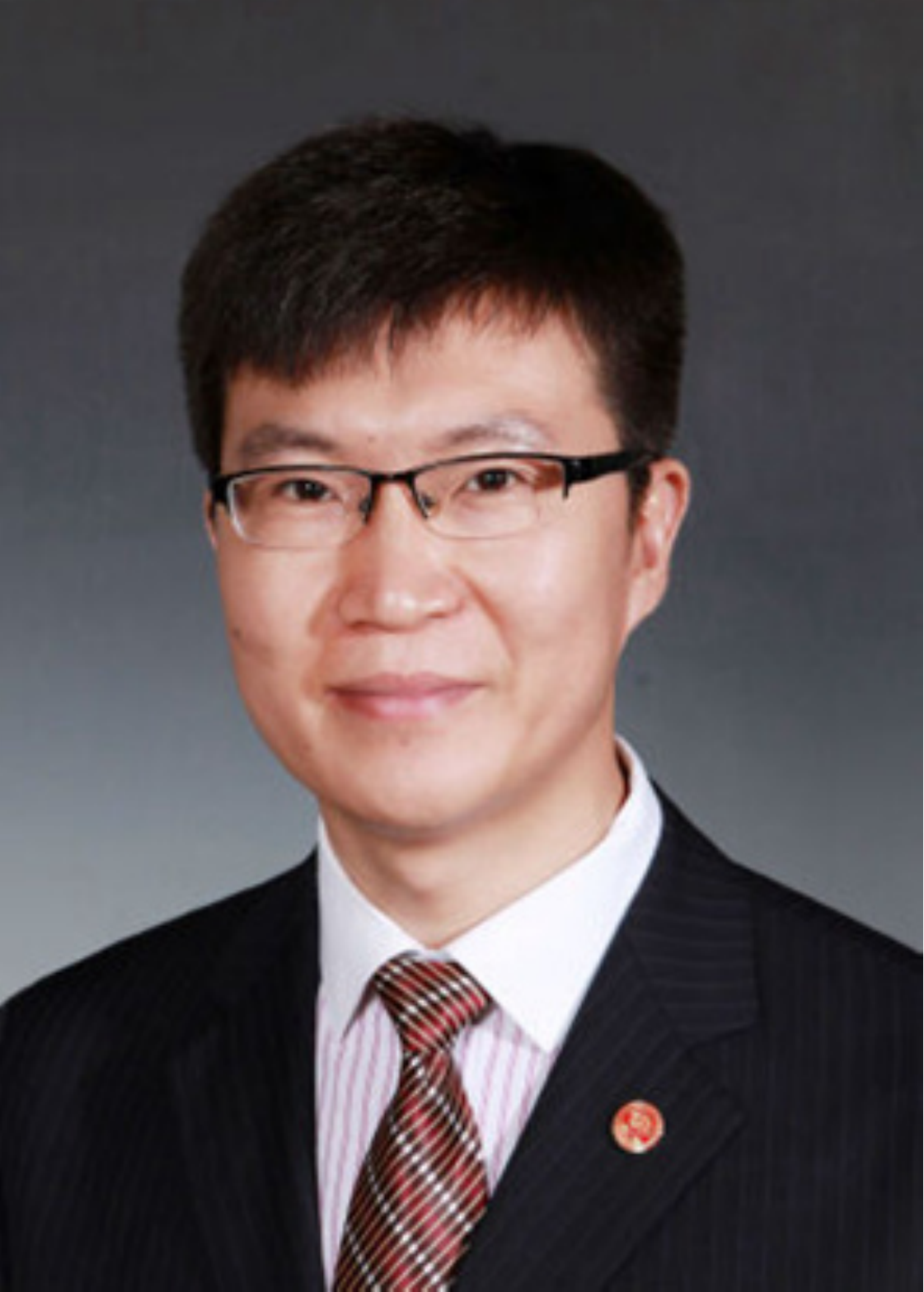}}]{Fusong Yuan}
received the B.A. degree in Dentistry from Weifang Medical University, Shandong, China, in 2008, and the M.E. degree and Ph.D. degree in Prosthodontics from Peking University, Beijing, China, in 2011 and 2014, respectively.

In 2014, he joined the Peking university school and hospital of Stomatology, as a dentist, teacher, and researcher. In 2021, he was promoted to associate chief physician. His current research interest includes the application research of robotic and femtosecond laser technology in stomatology.
\end{IEEEbiography}

\vspace{-1cm}
\begin{IEEEbiography}[{\includegraphics[width=1in,height=1.25in,clip,keepaspectratio]{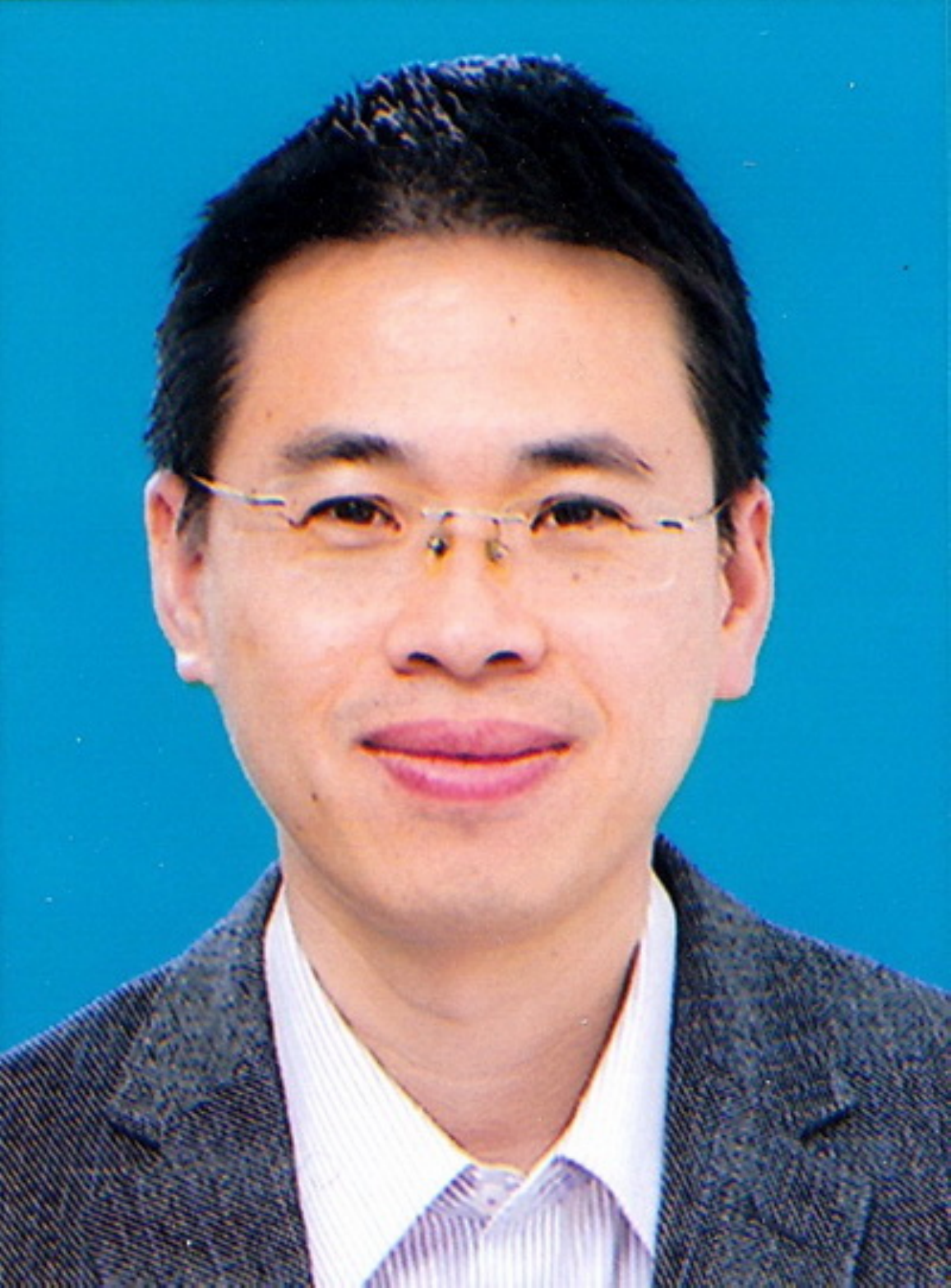}}]{Junzhi Yu (Fellow, IEEE)}
received the B.E. degree in safety engineering and the M.E. degree in precision instruments and mechanology from the
North University of China, Taiyuan, China, in 1998 and 2001, respectively, and the Ph.D. degree in control theory and control engineering from the Institute of Automation, Chinese Academy of Sciences, Beijing, China, in 2003.

From 2004 to 2006, he was a Postdoctoral Research Fellow with the Center for Systems and Control, Peking University, Beijing. In 2006, he was an Associate Professor with the Institute of Automation, Chinese Academy of Sciences, where he became a Full Professor in 2012. In 2018, he joined the College of Engineering, Peking University, as a Tenured Full Professor. His current research interests include intelligent robots, motion control, and intelligent mechatronic systems.
\end{IEEEbiography}

\end{document}